\documentclass[twoside]{article}

\usepackage[accepted]{aistats2025}
\usepackage[round]{natbib}
\usepackage{hyperref}       % hyperlinks

\usepackage{mathtools}
\usepackage{amssymb}
\usepackage{amsthm}
\usepackage{microtype}
\usepackage{bbm}
\usepackage{bbold}

\usepackage{graphicx}
\usepackage{booktabs} 
\usepackage{multirow}

\newtheorem{assumption}{Assumption}

\begin{document}

% \runningtitle{I use this title instead because the last one was very long}


\twocolumn[

\aistatstitle{HACSurv: A Hierarchical Copula-Based Approach for Survival Analysis with Dependent Competing Risks}

\aistatsauthor{ Xin Liu \And Weijia Zhang  \And  Min-Ling Zhang }

\aistatsaddress{ Southeast University \\xliu@seu.edu.cn \And  The University of Newcastle \\weijia.zhang@newcastle.edu.au \And Southeast University \\ zhangml@seu.edu.cn} ]
 % \aistatsaddress{ Southeast University \\School of Computer Science and Engineering \And  The University of Newcastle \And Southeast University \\School of Computer Science and Engineering } ]

% \aistatsauthor{ Xin Liu$^{1,2}$ \And Weijia Zhang$^3$  \And  Min-Ling Zhang$^{1,2}$ }

% \aistatsaddress{ 1. School of Computer Science and Engineering, Southeast University, China \\ 2. Key Laboratory of Computer Network and Information Integration (Southeast University),\\
% Ministry of Education, China \\ 3. School of Information and Physical Sciences, The University of Newcastle, Australia}]

\begin{abstract}
In survival analysis, subjects often face competing risks; for example, individuals with cancer may also suffer from heart disease or other illnesses, which can jointly influence the prognosis of risks and censoring. Traditional survival analysis methods often treat competing risks as independent and fail to accommodate the dependencies between different conditions. In this paper, we introduce HACSurv, a survival analysis method that learns Hierarchical Archimedean Copulas structures and cause-specific survival functions from data with competing risks. HACSurv employs a flexible dependency structure using hierarchical Archimedean copulas to represent the relationships between competing risks and censoring. By capturing the dependencies between risks and censoring, HACSurv improves the accuracy of survival predictions and offers insights into risk interactions. Experiments on synthetic dataset demonstrate that our method can accurately identify the complex dependency structure and precisely predict survival distributions, whereas the compared methods exhibit significant deviations between their predictions and the true distributions. Experiments on multiple real-world datasets also demonstrate that our method achieves better survival prediction compared to previous state-of-the-art methods.
\end{abstract}

\section{INTRODUCTION}
Survival analysis is a statistical methodology for predicting the time until an event of interest occurs. It plays a significant role in fields such as medicine~\citep{friedman2015medicine1,yeh2016developmentmedicine2}, reliability testing~\citep{pena2004modelsreliability}, and finance~\citep{caselli2021survivalfinace1,bosco2006factorsfinance2}. 
For example, in healthcare survival analysis has been utilized to predict the relapse/death time of cancer; in reliability engineering, it has been adopted to study the maintenance life of manufacturing equipment.
The objective of survival analysis is to estimate the probability of an event happening at a specific time and the event time, providing insights into the associated risk.

Compared to standard regression problems, the key challenge in survival analysis is the handling of censoring~\citep{wang2022survtrace,gharari2023copula,emura2018analysisofSurvival} and competing risks~\citep{lee2018deephit,li2023evaluating}. 
In healthcare, censoring occurs when a subject loses contact or experiences other risks before the event of interest. In the latter case, the situation is termed as \emph{competing risks} where patients may have multiple diseases simultaneously, but only the occurrence time of one disease can be observed. For example, a patient with cancer may also have other comorbidities, such as heart disease~\citep{li2023evaluating}. 

Uncovering the dependencies between competing risks through a data-driven approach is of great practical importance, as it not only enables more accurate survival predictions but also helps answer questions such as: ``Are individuals with arteriosclerosis more likely to die from pneumonia than those without a heart condition?'' raised by~\cite{tsiatis1975nonidentifiability}. 
Addressing such questions is essential because mutual influences among multiple diseases are common in real-world scenarios. Most existing methods for survival analysis under competing risks directly optimize the Cumulative Incidence Function (CIF)~\citep{lee2018deephit,nagpal2021deep,jeanselme2023neuralFG}; these methods neglect attention to the dependency structure or are based on the assumption of independent competing risks. Moreover, existing methods are all based on the independent censoring assumption, which is often violated in practice. Additionally, these CIF-based methods cannot provide predictions of the marginal distributions, which are also very important objectives in survival analysis.

To address these limitations, \emph{copulas} can be used as powerful statistical tools for modeling dependencies among random variables~\citep{okhrin2017CopulaeinHighDimensions}. Recently, researchers in the statistical community have used copulas to model the relationships between two competing events and censoring to obtain better survival estimates~\citep{li2023evaluating}. However, their methods rely on the Cox Proportional Hazards (CoxPH) assumption~\citep{cox1972COXPH}, which is a strong and often violated assumption in practice. In addition, their method requires users to specify the copula family, and a wrong copula type can increase the predictive bias. 

The recently introduced DCSurvival method~\citep{zhang2024DCSurvival}, which uses the ACNet~\citep{ling2020deepAC}, has demonstrated effectiveness in handling the dependencies in single-risk scenarios. However, using a single Archimedean copula can only characterize symmetric dependency relationships~\citep{okhrin2017CopulaeinHighDimensions}, and thus is not sufficient for competing risks as the relationships among risks and censoring are often flexible and asymmetric~\citep{li2023evaluating,li2019modelingCause,lo2020nested}. For example, a study on breast cancer survivors found that the dependency between the time to relapse/second cancer (RSC) and the time to cardiovascular disease (CVD), and their dependencies on informative censoring time are different~\citep{li2023evaluating,davis2014BreastSurvivor}.

In this paper, we introduce HACSurv, a survival analysis framework designed to learn hierarchical Archimedean copulas (HACs) and the marginal survival distributions of competing events and censoring from right-censored survival data. HACSurv uses HAC to flexibly model asymmetric dependency structures. To the best of our knowledge, HACSurv is the first data-driven survival analysis method that models the dependency among competing risks and censoring. Our contributions are summarized as follows:

\begin{itemize}
    \item We propose \textit{HACSurv}, a novel survival analysis method that captures the asymmetric dependency structure among competing risks and censoring using HAC. We also introduce HACSurv (Symmetry), a simplified version that employs a single Archimedean copula to model the dependency structure, which allows for end-to-end training.
    
    \item We revisit the survival prediction objective in the context of dependent competing risks and censoring. Unlike previous methods that rely solely on marginal survival functions for predictions~\citep{gharari2023copula,zhang2024DCSurvival}, our approach introduces a novel method to predict the conditional cause-specific cumulative incidence function (CIF) for dependent competing risks.
    
    \item Our experiments on synthetic datasets demonstrate that HACSurv significantly reduces bias in predicting the marginal survival distributions. Our methods achieve state-of-the-art results in survival outcome prediction on multiple real-world datasets. Furthermore, the copulas learned by HACSurv among competing events can potentially aid practitioners in better understanding the associations between diseases. Codebase: \url{https://github.com/Raymvp/HACSurv}.
\end{itemize}

\section{PRELIMINARIES}
\subsection{Survival Data and likelihood}
A sample in the survival dataset \(\mathcal{D} = \left\{\left(\boldsymbol{x}_i, t_i, e_i\right)\right\}_{i=1}^N\) typically consists of three components: (1) a \(D\)-dimensional covariate \(\boldsymbol{x}\), (2) the observed event time \(t\), and (3) an event indicator \(e\). 
For a survival dataset with \(K\) competing risks, the event indicator \(e\) ranges over \(\mathcal{K} = \{0, 1, \cdots, K\}\), where \(e = 0\) represents censoring. 
In competing events scenarios, we consider that only one event can be observed, and the observed time \(t_i\) is the minimum of the potential occurrence times of all events and censoring. In the single-risk case, \(e\) is a binary indicator. 

We extend the likelihood from ~\cite{gharari2023copula} and ~\cite{zhang2024DCSurvival} to the case with competing risks. For a sample \((\boldsymbol{x}, t, e)\) in survival data, the likelihood can be expressed as:

% \begin{equation}
% \mathcal{L} = \prod_{k=0}^K \left[\Pr(T_k = t, \{T_i > t\}_{i \neq k} | \boldsymbol{x})\right]^{1 \{e=k\}}
% \label{eq1}
% \end{equation}

\begin{equation}
\mathcal{L} = \prod_{k=0}^K \left[\Pr(T_k = t, \{T_i > t\}_{i \neq k} | \boldsymbol{x})\right]^{\mathbb{1}_{\{e=k\}}}
\label{eq1}
\end{equation}

where \(\mathbb{1}_{\{e=k\}}\) is an indicator function.

In this paper, we denote the marginal distributions for competing events or censoring time as \(S_{T_k \mid X}(t \mid \boldsymbol{x}) = \Pr(T_k > t \mid \boldsymbol{x})\). The corresponding density function can be written as \(f_{T_k \mid X}(t \mid \boldsymbol{x}) = -\frac{\partial S_{T_k \mid X}(t \mid \boldsymbol{x})}{\partial t}\).

If the events and censoring are assumed to be independent, Equation \ref{eq1} simplifies to:
\begin{equation}
\mathcal{L}_{\text{Indep}} = \prod_{k=0}^{K} \left[ f_k(t) \prod_{i \neq k} S_i(t) \right]^{\mathbb{1}_{\{e=k\}}}.
\end{equation}

% However, in reality, events and censoring are often dependent \citep{li2019modelingCause}. In this paper, we use \emph{copulas} to model their dependencies, under the mild assumption that the copula does not depend on the covariates \(\boldsymbol{x}\).
However, in reality, events and censoring are often dependent \citep{li2019modelingCause}. In this paper, we use \emph{copulas} to model their dependencies under the following assumption:
\begin{assumption}
We assume that the copula $C$ used to describe the dependency structure among competing risks and censoring does not depend on covariates \(\boldsymbol{x}\).
\end{assumption}
In some medical contexts, such as gender differences affecting the dependency between diseases A and B, the copula might depend on \(\boldsymbol{x}\), but such cases may not be common or the variation in the copula may not be significant. We leave research on relaxing this assumption for future endeavours.

\subsection{Copulas and Archimedean Copulas}\label{sec2.2}
Copula is a powerful statistical tool for modeling the dependence structure between multivariate random variables, which can be viewed as a multivariate cumulative distribution function (CDF) with uniform \([0, 1]\) margins. Sklar's theorem ensures that any \(d\)-dimensional continuous CDF can be uniquely expressed as a composition of its univariate margins and a copula.
\newtheorem{theorem}{Theorem}
\begin{theorem}[Sklar's theorem]
For a \(d\)-variate cumulative distribution function \(F\), with \(j\)-th univariate margin \(F_j\), the copula associated with \(F\) is a cumulative distribution function \(C:[0,1]^d \rightarrow [0,1]\) with \(U(0,1)\) margins satisfying:
\begin{equation}
F(x_1, \cdots, x_d) = C\left(F_1\left(x_1\right), \cdots, F_d\left(x_d\right)\right), 
\end{equation}
where \((x_1, \cdots, x_d) \in \mathbb{R}^d \). If \(F\) is continuous, then \(C\) is unique.
\end{theorem}
Due to its flexibility, non-parametric copulas are difficult to characterize. 

\textbf{Archimedean Copulas} \quad
% \paragraph{Archimedean Copulas}
Most recent works focus on Archimedean copulas \citep{ling2020deepAC,ng2021generative,zhang2024DCSurvival,gharari2023copula}, a widely used family of copulas that can represent different tail dependencies using a one-dimensional generator:
\begin{equation} \label{eq:4}
 C(\mathbf{u}) = \varphi\left(\varphi^{-1}\left(u_1\right) + \cdots + \varphi^{-1}\left(u_d\right)\right),
\end{equation}
where \(\varphi: [0, \infty) \rightarrow [0, 1]\) is the generator of the Archimedean copula. 
To ensure that $C$ is a valid copula, the generator must be completely monotone, i.e., \( (-1)^k \varphi^{(k)} \geq 0 \) for all \(k \in \{0, 1, 2, \cdots\} \).
\subsection{Hierarchical Archimedean Copulas}
A significant limitation of Archimedean copulas, which are defined by a single generation, is their inability to model asymmetric dependency structures~\citep{okhrin2017CopulaeinHighDimensions}. However, in real-world scenarios, the dependencies among events and censoring are rarely symmetric~\citep{li2019modelingCause}. For example, the dependence between the time to relapse or a second cancer and cardiovascular disease may differ significantly from their dependence on censoring~\citep{li2023evaluating}.
% Directly modeling the dependencies between event times and censoring using standard Archimedean copulas can therefore introduce bias.

\begin{figure*}
    \centering
    \includegraphics[width=0.9\linewidth]{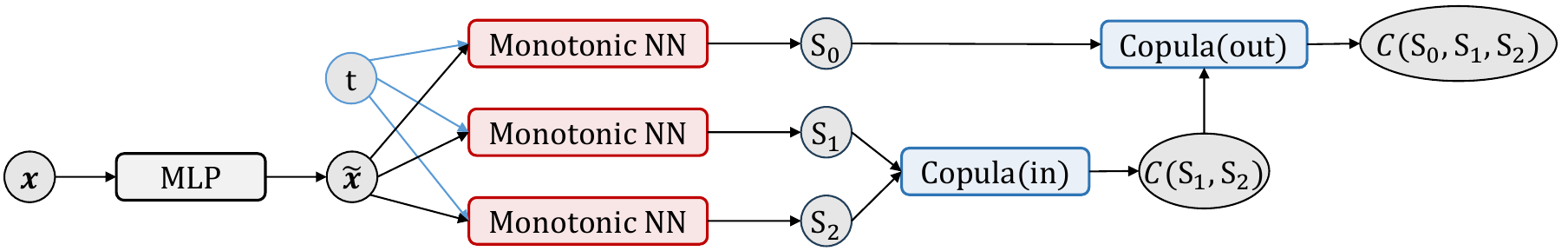}
    \caption{Overview of our HACSurv for the two competing risks scenario. We abbreviate \(S_{T_k \mid X}\) as \(S_k\), for \(k = 0, 1, 2\). The copula (out) can be represented by the outer generator \(\varphi_0\). The inner generator \(\varphi_1\) corresponding to copula (in) is constructed from the Laplace exponent \(\psi_1\) and the outer generator \(\varphi_0\).}
    \label{fig:1}
\end{figure*}

Hierarchical Archimedean Copulas (HACs) address these limitations by allowing more complex and flexible dependency structures~\citep{joe1997multivariate}.To illustrate the idea of HAC, we consider a three-dimensional example as shown in Figure~\ref{fig:1}. This approach can be easily extended to higher-dimensional cases with more inner copulas. The HAC for a three-dimensional scenario is expressed as:
%删去这个图，增加空间
% \begin{figure}
%     \centering
%     \includegraphics[width=0.35\linewidth]{figures/2risksExample.pdf}
%     \caption{An HAC structure for two competing risks. }
%     \label{fig:1}
% \end{figure}

% \begin{align}
% C\left(u_1, u_2, u_3\right) &= \varphi_0\left(\varphi_0^{-1}\left(u_1\right) + \right. \\
% &\quad \left. \varphi_0^{-1} \circ \varphi_1\left(\varphi_1^{-1}\left(u_2\right) + \varphi_1^{-1}\left(u_3\right)\right)\right)
% \end{align}

\begin{equation}
\begin{aligned}
C\left(u_1, u_2, u_3\right) &= \varphi_0\left(\varphi_0^{-1}\left(u_1\right) + \right. \\
&\quad \left. \varphi_0^{-1} \circ \varphi_1\left(\varphi_1^{-1}\left(u_2\right) + \varphi_1^{-1}\left(u_3\right)\right)\right).
\end{aligned}
\end{equation}

For an HAC to be a valid copula, the sufficient nesting conditions~\citep{joe1997multivariate,mcneil2008sampling} are required, which state that:

\begin{itemize}
    \item \(\varphi_j\) for all \(j \in \{0, 1, \ldots, j\}\) are completely monotone,
    \item \(\left(\varphi_0^{-1} \circ \varphi_j\right)^{\prime}\) for \(j \in \{1, \ldots, j\}\) are completely monotone.
\end{itemize}

The latter criterion, i.e., \(\left(\varphi_0^{-1} \circ \varphi_j\right)^{\prime}\) is completely monotone, is discussed in detail in~\citep{hering2010constructing}. In brief, for a given outer generator \(\varphi_0\), the inner generator \(\varphi_j\) can be constructed using the Laplace exponent \(\psi_j\) of a Lévy subordinator and the outer generator \(\varphi_0\), such that \(\varphi_j(x) = (\varphi_0 \circ \psi_j)(x)\). The Laplace exponent \(\psi_j\) has the following expression:

\begin{align}
\psi_j(x) &= \mu_j x + \beta_j \left( 1 - \varphi_{M_j}(x) \right) \label{eq:firstLineJ} \\
          &= \mu_j x + \beta_j \left( 1 - \int_0^{\infty} e^{-xs} dF_{M_j}(s) \right) \label{eq:7},
\end{align}

where \(M_1\) is a positive random variable with Laplace transform \(\varphi_{M_1}\), and \(\mu_1 > 0\) and \(\beta_1 > 0\). The derivation of \(\psi_1\) is rather complex, so we only present the final form of \(\psi_1\) here. The detailed mathematical derivation is provided in the supplementary material following~\cite{hering2010constructing}.

\section{METHODS}
% The goal of HACSurv is to use hierarchical Archimedean copula to model the asymmetric dependency structure among competing risks and censoring, and learn the corresponding cause-specific marginal survival functions with neural network. 
In order to reduce estimation bias caused by independent or symmetric dependencies, HACSurv utilizes hierarchical Archimedean copula to model the asymmetric dependency structure among competing risks and censoring, and learns the corresponding cause-specific marginal survival functions with neural network. 
Figure \ref{fig:1} illustrates the model architecture of HACSurv in the setting of 2 competing risks. 

In this section, we first derive the likelihood under dependent competing risks and censoring, and introduce the first method proposed in this paper, which uses a single Archimedean copula to characterize the overall dependency of the competing risks. 
% Although this method cannot fully capture the complex asymmetric dependency structure, its advantage is that both \(\varphi\) and the marginal survival functions can be trained end-to-end.
Next, we present how neural networks can be used to model HAC and the marginal survival distributions. The final part of this section explores how to compute the conditional cause-specific cumulative incidence function (CIF) using HACSurv in the presence of dependencies, which is an alternative survival prediction method to directly use the marginal survival distribution.

\subsection{Survival Likelihood with Dependent Competing Risks}
We begin by extending the survival likelihood with dependent censoring described in~\cite{zhang2024DCSurvival} to the HAC-based survival likelihood with dependent competing risks. 
Using Sklar's theorem, we obtain the joint probability function for \(K\) competing risks:
\begin{equation} \label{eq8}
\begin{aligned}
\Pr(T_0 > t&, \cdots, T_K > t \mid \boldsymbol{x}) = \\&\mathcal{C}\left(S_{T_0 \mid X}(t \mid \boldsymbol{x}), \cdots, S_{T_K \mid X}(t \mid \boldsymbol{x})\right).
\end{aligned}
\end{equation}
%原来的公式双栏实在放不下。
% Incorporating the above equation into the survival likelihood in Equation~\ref{eq1}, we can get:
% \begin{equation} \label{eq9}
% \begin{aligned}
% \mathcal{L} &= \prod_{k=0}^{K} \left\{ -\left.\frac{\partial}{\partial u_k} \operatorname{Pr}\left(T_k > u_k, \left\{T_i > t\right\}_{i \neq k} \mid X = \boldsymbol{x}\right) \right|_{u_k=t} \right\}^{1\{e=k\}} \\
% &= \prod_{k=0}^{K} \left\{ f_{T_k|X}(t|\boldsymbol{x}) \frac{\partial}{\partial u_k} C(u_0, \cdots, u_K) \left|_{\begin{smallmatrix} u_k = S_{T_k|X}(t|\boldsymbol{x})\end{smallmatrix}} \right. \right\}^{1\{e =k\}}.
% \end{aligned}
% \end{equation}

Incorporating the above equation into the survival likelihood in Equation~\ref{eq1} (we omit covariates $\boldsymbol{x}$ for brevity), we can get:
% \begin{equation} \label{eq9}
% \mathcal{L} = \prod_{k=0}^{K} \left\{ f_{T_k}(t) \frac{\partial}{\partial u_k} C(u_0, \dots, u_K) \left|_{\begin{smallmatrix} u_k = S_{T_k}(t)\end{smallmatrix}} \right. \right\}^{1\{e =k\}}
% \end{equation}
\begin{equation} \label{eq9}
\mathcal{L} = \prod_{k=0}^{K} \left\{ f_{T_k}(t) \frac{\partial}{\partial u_k} C(u_0, \dots, u_K) \right\}^{\mathbb{1}_{\{e=k\}}},
\end{equation}
where each \( u_i = S_{T_i}(t) \) for all \( i = 0, \dots, K \).
% We provide the full derivation in the supplementary material.
A simple approach to represent $C(u_0, \dots, u_K)$ is by using a single Archimedean copula according to Equation~\ref{eq:4}. This survival analysis method based on a single copula is referred to in this paper as HACSurv (Symmetry), which is an extension of DCSurvival~\citep{zhang2024DCSurvival} in the competing risks setting. Although it cannot capture complex asymmetric structures, it remains a practical approach. Later experiments demonstrate its superiority over the independent copula. Furthermore, a key advantage of HACSurv(Symmetry) is that it can be trained in an end-to-end manner.

A more flexible modeling approach is to represent $C(u_0, \dots, u_K)$ using a hierarchical Archimedean copula where \(\frac{\partial}{\partial u_k} C(u_0, \cdots, u_K)\) can be calculated based on the HAC structure and the chain rule of derivatives. For the example shown in Figure~\ref{fig:1}, when $k = 1$, this can be specifically expressed as: 
\begin{equation}
\begin{aligned}
\frac{\partial}{\partial u_1} &C(u_0, u_1, u_2) = \\&
\frac{\partial C(u_0, u_1, u_2)}{\partial C_{in}(u_1, u_2)} \cdot 
\frac{\partial C_{in}(u_1, u_2)}{\partial u_1} \left|_{\begin{smallmatrix} 
u_0 = S_{T_0}(t) \\ 
u_1 = S_{T_1}(t) \\ 
u_2 = S_{T_2}(t) 
\end{smallmatrix}} \right.,
\end{aligned}
\end{equation}
where \(C_{in} = \varphi_1\left(\varphi_1^{-1}\left(u_1\right) + \varphi_1^{-1}\left(u_2\right)\right)\).
Under this HAC structure, \(C_{in}\) can capture diverse forms of dependency between competing events. Moreover, the outer copula can describe the dependencies between competing events and censoring.
\subsection{Learning Archimedean Copula from Survival Data}

As discussed in Section~\ref{sec2.2}, we need a completely monotone generator \(\varphi\) to construct a valid Archimedean copula. The Bernstein-Widder characterization theorem~\citep{bernstein1929fonctions,widder2015laplaceTrans}, states that any completely monotone function can be characterized by the Laplace transform of a positive random variable:
\begin{theorem}[Bernstein-Widder]
A function \(\varphi\) is completely monotone and \(\varphi(0) = 1\) if and only if \(\varphi\) is the Laplace transform of a positive random variable. Specifically, \(\varphi(x)\) can be represented as:
\begin{equation}
\varphi(x) = \int_{0}^{\infty} e^{-xs} \, dF_M(s),
\end{equation}
where \(M > 0\) is a positive random variable with the Laplace transform \(\varphi\).
\end{theorem}
We represent the generator \(\varphi\) of Archimedean copula using the method proposed by~\cite{ng2021generative}.
Specifically, we define a generative neural network \(G(\cdot; \theta)\) parameterized by \(\theta\). We let \(M\) be the output of this neural network such that samples \(M \sim F_M\) are computed as \(M = G(\epsilon; \theta)\), where \(\epsilon\) is a source of randomness and serve as the input to the neural network. We then approximate the Laplace transform \(\varphi(x)\) by its empirical version, using \(L\) samples of \(M\) from \(G(\cdot; \theta)\):
\begin{equation} \label{eq:12}
\varphi(x) = \int_{0}^{\infty} e^{-xs} \, dF_M(s) = \mathbb{E}[e^{-Mx}] \approx \frac{1}{L} \sum_{l=1}^{L} e^{-M_lx},
\end{equation}
To compute Equation~\ref{eq:4}, we need to compute the inverse of \(\varphi(x)\). Following previous works \citep{ling2020deepAC,ng2021generative}, we employ Newton's method to achieve this. Additionally, by differentiating with respect to \(u_k = S_k(t)\), we can derive \(\frac{\partial}{\partial u_k} C(u_0, \ldots, u_K)\) in Equation~\ref{eq9}. During the optimization of the survival likelihood, the parameters of the generative neural network \(G(\cdot; \theta)\) are updated via gradient descent.

\subsection{Learning HAC with the Re-generation Trick}
\subsubsection{Determine the Structure of HAC} \label{section3.3.1}
The ideal Hierarchical Archimedean Copula (HAC) structure should closely resemble the true dependency structure among competing events and censoring. For survival data with only one event or censoring observation time, it is theoretically not guaranteed to determine the true marginal distribution and copula without making any assumptions~\citep{tsiatis1975nonidentifiability}. %考虑一下放在这里还是放在后面专门讨论可识别性 
However, this unidentifiability does not make the estimation pointless. As we will demonstrate in the experiments, accounting for the dependency among events substantially reduces estimation bias, despite the unidentifiable results.

% In this paper, we first capture the copula between any two pairs of competing events and censoring. The approach is similar to DCSurvival proposed by \cite{zhang2024DCSurvival} for determining the copula between events and censoring, but we replace ACNet \cite{ling2020deepAC} with Gen-AC \cite{ng2021generative} to represent the generator \(\varphi\) which enhances computational efficiency. When determining the HAC structure, one principle is that the dependency strength of the inner copulas should be stronger than that of the outer copulas \cite{okhrin2017CopulaeinHighDimensions}. For the case with two competing risks shown in Figure~\ref{fig:1}, where there is one inner and one outer copula, we recommend using the copula with the weakest dependency as the outer copula and the one with the strongest dependency as the inner copula. As the number of competing risks increases, the structure of the HAC can also be flexibly determined based on the dataset, and we provide additional examples with three and five competing risks in the experimental section.

In this paper, we first capture the copulas between each pair of competing events and censoring. The approach is similar to DCSurvival proposed by \cite{zhang2024DCSurvival} for determining the copula between events and censoring, but we replace ACNet \citep{ling2020deepAC} with Gen-AC \citep{ng2021generative} to represent the generator \(\varphi\) which enhances computational efficiency. When determining the HAC structure, we follow the principle that the dependency strength of inner (lower-level) copulas should be stronger than that of outer (higher-level) copulas~\citep{okhrin2017CopulaeinHighDimensions,li2019modelingCause}. Thus, the basic idea for constructing the HAC structure is to group the competing events and censoring based on their dependencies: strong dependencies within groups are modeled by lower-level inner copulas, while weaker dependencies between groups are captured by higher-level outer copulas. In the experimental section, we provide examples with 2, 3, and 5 competing risks to specifically demonstrate how to determine the HAC structure.

\subsubsection{Training inner generator with Re-generation Trick}

For the inner generator \( \varphi_j \), we define it using the composition \( \varphi_j(x) = (\varphi_0 \circ \psi_j)(x) \), following the approach proposed in~\cite{ng2021generative}. We configure the parameters \( \mu \) and \( \beta \) in Equation~\ref{eq:7} as trainable parameters, applying an \( \exp(\cdot) \) output activation to ensure positivity. \( M_j \) is set as the output from \( G(\cdot, \theta_j) \) with parameters \( \theta_j \) and \( \exp(\cdot) \) output activation. We proceed by computing the Laplace transform \(\varphi_{M_j}\) as outlined in Equation~\ref{eq:12}.
% We can approximate its derivatives \(\varphi^{(k)}_{M_j}\) using the empirical version of the Laplace transform's derivatives:
% \begin{equation} \label{er:13}
% \varphi^{(k)}(x) = \mathbb{E}_M\left[(-M)^k e^{-Mx}\right] \approx \frac{1}{L} \sum_{l=1}^L (-M_l)^k e^{-M_lx}.
% \end{equation}

Although \cite{ng2021generative} presents a method for training the inner generator with fully observed data, their approach does not apply to partially observed survival data. To address competing risks, this paper introduces a two-stage training strategy tailored for survival data:
\begin{enumerate}
    \item The first stage involves generating a collection of bivariate data points, denoted as \(\mathbf{U}\), each represented by coordinates \((U_1, U_2)\sim C\). These data points are sampled from the selected copula, identified in Section~\ref{section3.3.1} as the inner copula from among several candidates. This dataset is then utilized to train the inner generator \( \varphi_j \) of the HAC.
    \item The second stage focuses on training \( \varphi_j \) using Maximum Likelihood Estimation (MLE) based on the dataset generated in the first stage. 
    % To compute the MLE, we specifically require the second-order derivatives of the Laplace transform, \(\varphi_{j}''\). 
    To compute the MLE, we specifically require the second-order derivatives of the inner generator, \(\varphi_{j}''\). Just as in~\cite{ling2020deepAC}, we utilize PyTorch~\citep{paszke2017automatic} for automatic differentiation.

\end{enumerate}
During this process, we keep the outer generator \( \varphi_0 \) fixed, following~\cite{ng2021generative}

\subsubsection{Using a Specified HAC}
In addition to the data-driven approach for determining the parameters of the HAC described above, some practitioners might be interested in specifying a known copula according to their prior knowledge. Based on our HACSurv framework, using a specified HAC for survival analysis is also feasible. Specifically, one can either employ the state-of-the-art HACopula Toolbox~\citep{gorecki2017structure} to generate multidimensional samples $\mathbf{U} \sim C$ from the specified HAC, or generate a set of bivariate samples for each specified outer and inner copula. Then, using these samples, the outer copula parameters can be estimated via maximum likelihood estimation. For the inner copula, the training method described in the second stage above can be applied directly.

\subsection{Survival Analysis via HACSurv}
\subsubsection{Learning Marginal Distributions} 
Once the HAC that describes the dependency structure among competing events and censoring is obtained, our goal is to learn their survival distributions. First, we set up a shared embedding network to extract features from the covariates $\boldsymbol{x}$. We use monotonic neural density estimators (MONDE)~\citep{chilinski2020neural} to model \(S_{k \mid X}\)for all \(k\) from 0 to \(K\) following DCSurvival~\citep{zhang2024DCSurvival} and SuMo-net~\citep{rindt2022survival}. MONDE consists of two parts: the first part is a fully connected network that processes the covariates, and the processed covariates along with the time input \(t\) are then concatenated and fed into the second part of the network. In this second part, all weights of the successor nodes of the time input \(t\) are non-negative. The final layer output of the MONDE network is transformed into the survival function through the sigmoid function. The corresponding density functions \(f_{k \mid X}\) can be computed using PyTorch's automatic differentiation~\citep{paszke2017automatic}.
%考虑缩减monde的描述

After both the HAC and the marginals are instantiated, Equation~\ref{eq9} can be computed based on the HAC structure and the chain rule of differentiation, and the parameters of the marginals can be optimized using stochastic gradient descent. It is worth noting that during the estimation of the final parameters of the marginals, we freeze the parameters of the HAC. 
%下面的也可删减
% This is done for the following reasons:
% \begin{enumerate}
%     \item We have already established the reasonable structure and parameters of the HAC in previous steps.
%     \item Under insufficient conditions, the competing risk model is non-identifiable~\cite{tsiatis1975nonidentifiability,wang2012nonidentifiability}, making it impossible for the model to find reasonable copula parameters.
% \end{enumerate}
% Empirically, after unfreezing the HAC parameters, we found that the model did not learn a meaningful copula.

\subsubsection{Predicting Cause-Specific CIF} 
Previous copula-based survival analysis methods ~\citep{zhang2024DCSurvival,gharari2023copula} employ marginal survival distribution for prediction.
However, as the dependency within the true copula of event and censoring increases, \citet{gharari2023copula} have shown that the Integrated Brier Score (IBS) obtained from the ground-truth marginal survival function also increases, and thus IBS is not suitable for dependent censoring. 

Moreover, we posit that the observed survival data represent the joint probability of the event \( e_i \) occurring at time \( t_i \) and other events not occurring at that time rather than simply marginal probability. To better account for the dependencies among competing events and censoring, we argue that predicting the conditional Cause-Specific Cumulative Incidence Function (CIF) offers a superior approach compared to predicting the marginal.  To this end, we extend the original CIF~\citep{fine1999proportional,lee2018deephit} to accommodate scenarios with dependent competing risks and censoring:
% \begin{align}
% F_{k^*}\left(t^* \mid \boldsymbol{x}^*\right) &= \Pr(T_{k^*} < t^* \mid \{T_i > t^*\}_{i \neq k^*}, \boldsymbol{x}^*) \\
% &= 1 - \frac{C(\{u_i\}_{i \neq k^*})}{C(\{u_i\}_{i = 0, \dots, K})}
% \end{align}

\begin{align}
F_{k^*}\left(t^* \mid \boldsymbol{x}^*\right) &= \Pr(T_{k^*} < t^* \mid \{T_i > t^*\}_{i \neq k^*}, \boldsymbol{x}^*) \\
&= 1 - \frac{C(\{u_i\}_{i = 0, \dots, K})}{C(\{u_i\}_{i \neq k^*})},
\end{align}
where each \(u_i = S_{T_i \mid X}(t^* \mid \boldsymbol{x}^*)\) for all \(i = 0, \dots, K\).
% However, due to the complexity of survival analysis problems, this paper retains both the CIF and Marginal methods as two prediction approaches.

\section{DISCUSSION ON IDENTIFIABILITY}
% \label{sec-unidentifiability}
% \citet{tsiatis1975nonidentifiability} has shown that for survival data generated from a ground-truth joint probability distribution (Equation~\ref{eq8}) corresponding to the dependent competing risks and censoring in survival data , there exists an independent copula \(C_{\text{Indep}}\) and corresponding marginal survival functions \(G_{T_k \mid X}(t \mid x)\) for all \(k \in \{0, \ldots, K\}\)  that can also generate the observed data. 

\citet{tsiatis1975nonidentifiability} has shown that, even when survival data are generated from a ground-truth joint probability distribution (Equation~\ref{eq8}) representing dependent competing risks and censoring, there exists an independent copula \(C_{\text{Indep}}\) and corresponding marginal survival functions \(G_{T_k \mid X}(t \mid x)\) for all \(k \in \{0, \ldots, K\}\) that can produce the same observed data.
In fact, there are infinite combinations of copulas and corresponding marginal distributions that can represent this joint probability. Most survival methods that do not rely on copulas only optimise the cause-specific cumulative incidence function (CIF) corresponding to this joint probability, without considering the marginal distribution or assuming competing risks and censoring are mutually independent~\citep{lee2018deephit,nagpal2021deep,danks2022DeSurv,jeanselme2023neuralFG}.
%是否需要表示出来他们的CIF？
% \begin{align}
% H_{k^*}\left(t^* \mid \boldsymbol{x}^*\right) &= \Pr(T_{k^*} < t^*, \{T_i > t^*\}_{i \neq k^*} \mid \boldsymbol{x}^*) 
% \end{align}

However, the unidentifiability result does not render efforts to estimate the copula among competing events futile.
The greater the difference between the estimated copula and the ground truth copula, the greater the disparity between the obtained and true marginal distributions. 
In other words, if the competing risks are dependent, existing methods will exhibit significant bias since they inherently assume an independent copula.
Although it is impossible to \emph{uniquely identify} the ground-truth copula, our method can \emph{closely approximate} it given the available data.
% maybe change this a little according to the actual experiment results, but follow the general logic
As we demonstrated in the experiments, under metrics that compared the estimated survival marginals to the ground truth distributions (Survival-l1)~\citep{gharari2023copula}, HACSurv significantly reduces estimation biases when compared to methods that assume independence; under metrics that do not require access to the ground truth (e.g., \( C^{td} \)-index~\citep{antolini2005cindex} and IBS~\citep{graf1999IBS}), HACSurv also performs significantly better.

% Although there is no comprehensive theoretical identifiability guarantee, experiments have demonstrated that our approach, compared to existing state-of-the-art survival methods, achieves significantly better estimates of the marginal survival distribution and survival outcomes.

\section{EXPERIMENTS}
In this section, we evaluate HACSurv's ability to capture the dependencies among competing risks and censoring, as well as the model's predictive ability for the marginal survival distribution and survival outcomes. Since the true dependencies are not known in the real world, we first construct a synthetic dataset with three competing risks based on a known HAC. In addition, we further compare against baselines using three real-world datasets to evaluate HACSurv's ability to predict patient survival time. We pick the following baselines for comparison: DeepHit~\citep{lee2018deephit}, DSM~\citep{nagpal2021deep}, DeSurv~\citep{danks2022DeSurv}, and NeuralFG~\citep{jeanselme2023neuralFG}. We also present the results of HACSurv with an independent copula and a single Archimedean copula. 

Due to the complexity of the survival analysis problem, we chose whether to use the marginal survival function or CIF for prediction based on the average \( C^{td} \)-index of all risks in the validation set. 
\begin{figure}
    \centering
    \includegraphics[width=1\linewidth]{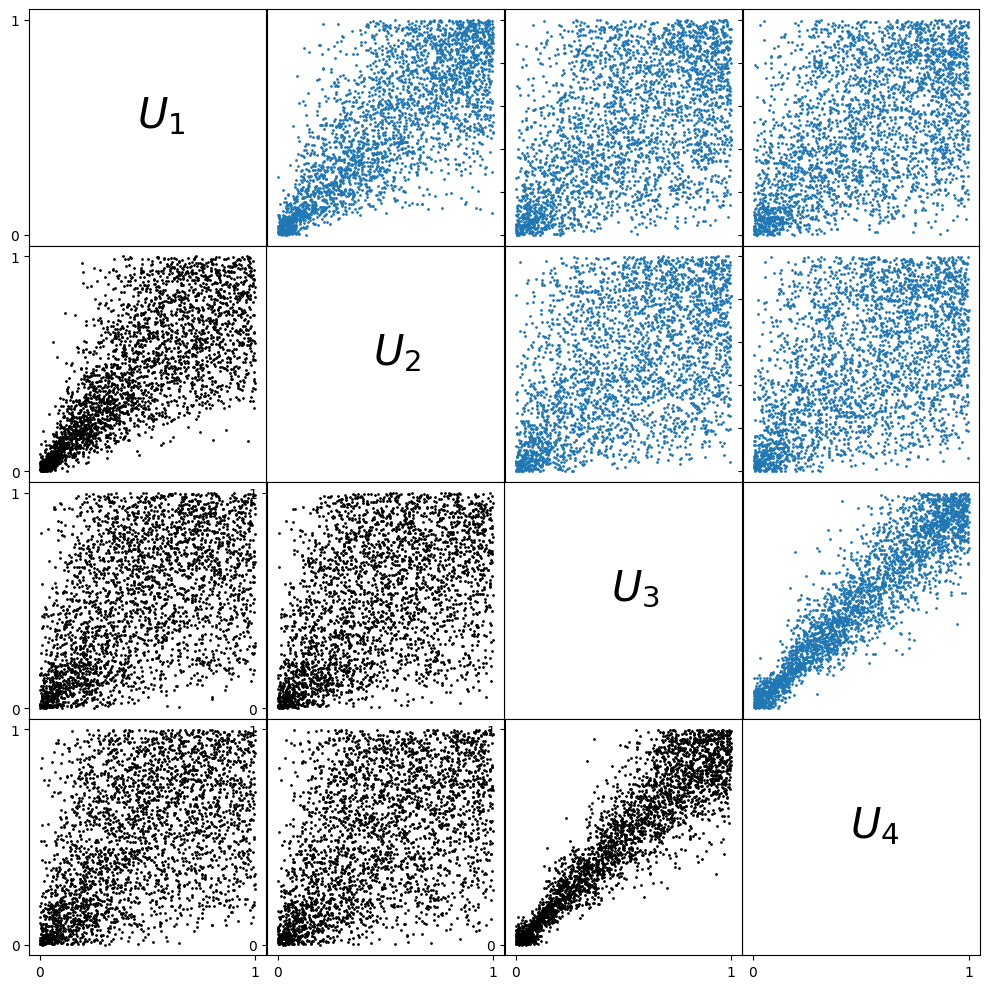}
    \caption{The blue samples are generated from the HAC learned on the synthetic dataset. The samples drawn from the ground truth copulas are in black. The results are presented in a mirrored format.}
    \label{fig:2}
\end{figure}

\paragraph{Synthetic Dataset} Following the approach proposed by~\cite{gharari2023copula} for constructing a single-risk synthetic dataset with dependent censoring, we develop a synthetic dataset with three competing risks. The marginal distributions of competing events and censoring are modeled using a CoxPH model where the events and censoring risks are specified by Weibull distributions where the hazards are linear functions of covariates. The dependency structure is characterized by a known four-variate hierarchical Archimedean copula \(C_{\varphi_0}\left( C_{\varphi_1}\left( u_1, u_2 \right), C_{\varphi_2}\left( u_3, u_4 \right) \right)\). The samples of the known copula are generated using the HACopula Toolbox~\citep{gorecki2017structure}. Specifically, \(C_{\varphi_0}, C_{\varphi_1}, C_{\varphi_2}\) are Clayton copulas with their \(\theta\) parameters set to 1, 3, and 8, respectively. 

\paragraph{Real-World Datasets} 
\textit{Framingham}: This is a cohort study gathering 18 longitudinal measurements on male patients over 20 years~\citep{kannel1979Framingham}. We consider death from cardiovascular disease (CVD) as Risk~2, and death from other causes as Risk~1.

\textit{SEER}: The Surveillance, Epidemiology, and End Results Program (SEER) dataset~\citep{howlader2010improved} is an authoritative database in the U.S. that provides survival information for breast cancer patients. We select information on breast cancer patients during 2000--2020. Among a total of 113,561 patients, the percentages of censored, died due to cardiovascular disease (Risk~1), and died due to breast cancer (Risk~2) are approximately 70.9\%, 6\%, and 23.1\%, respectively.

\textit{MIMIC-III}: We extract survival data of 2,279 patients from the MIMIC-III database~\citep{tang2021mimic}. Among them, 1,353 patients (59.37\%) are right-censored; 517 patients (22.68\%) died of sepsis (Risk~1); 65 patients (2.85\%) died of cerebral hemorrhage (Risk~2); 238 patients (10.44\%) died of acute respiratory failure (Risk~3); 62 patients (2.72\%) died of subendocardial acute myocardial infarction (Risk~4); and 44 patients (1.93\%) died of pneumonia (Risk~5).

% \begin{figure*}
%     \centering
%     \includegraphics[width=0.8\linewidth]{HAC_structure.pdf}
%     \caption{Enter Caption}
%     \label{fig:enter-label}
% \end{figure*}

% \paragraph{Framingham} This is a cohort study gathering 18 longitudinal measurements on male patients over 20 years. We consider death from cardiovascular disease (CVD) as Risk2, and death from other causes as Risk1.
% \paragraph{SEER} The Surveillance, Epidemiology, and End Results Program (SEER) dataset is an authoritative database in the U.S. that provides survival information for breast cancer patients. We selected information on breast cancer patients during 2000--2020. Among a total of 113,561 patients, the percentages of censored, died due to cardiovascular disease (Risk 1), and died due to breast cancer (Risk 2) were approximately 70.9\%, 6\%, and 23.1\%, respectively.

% \paragraph{MIMIC-III} We extracted survival data of 2,279 patients from the MIMIC-III database. Among them, 1,353 patients (59.37\%) were right-censored; 517 patients (22.68\%) died of sepsis (Risk1); 65 patients (2.85\%) died of cerebral hemorrhage (Risk2); 238 patients (10.44\%) died of acute respiratory failure (Risk3); 62 patients (2.72\%) died of subendocardial acute myocardial infarction (Risk4); and 44 patients (1.93\%) died of pneumonia (Risk5).

\begin{figure}
    \centering
    \includegraphics[width=1\linewidth]{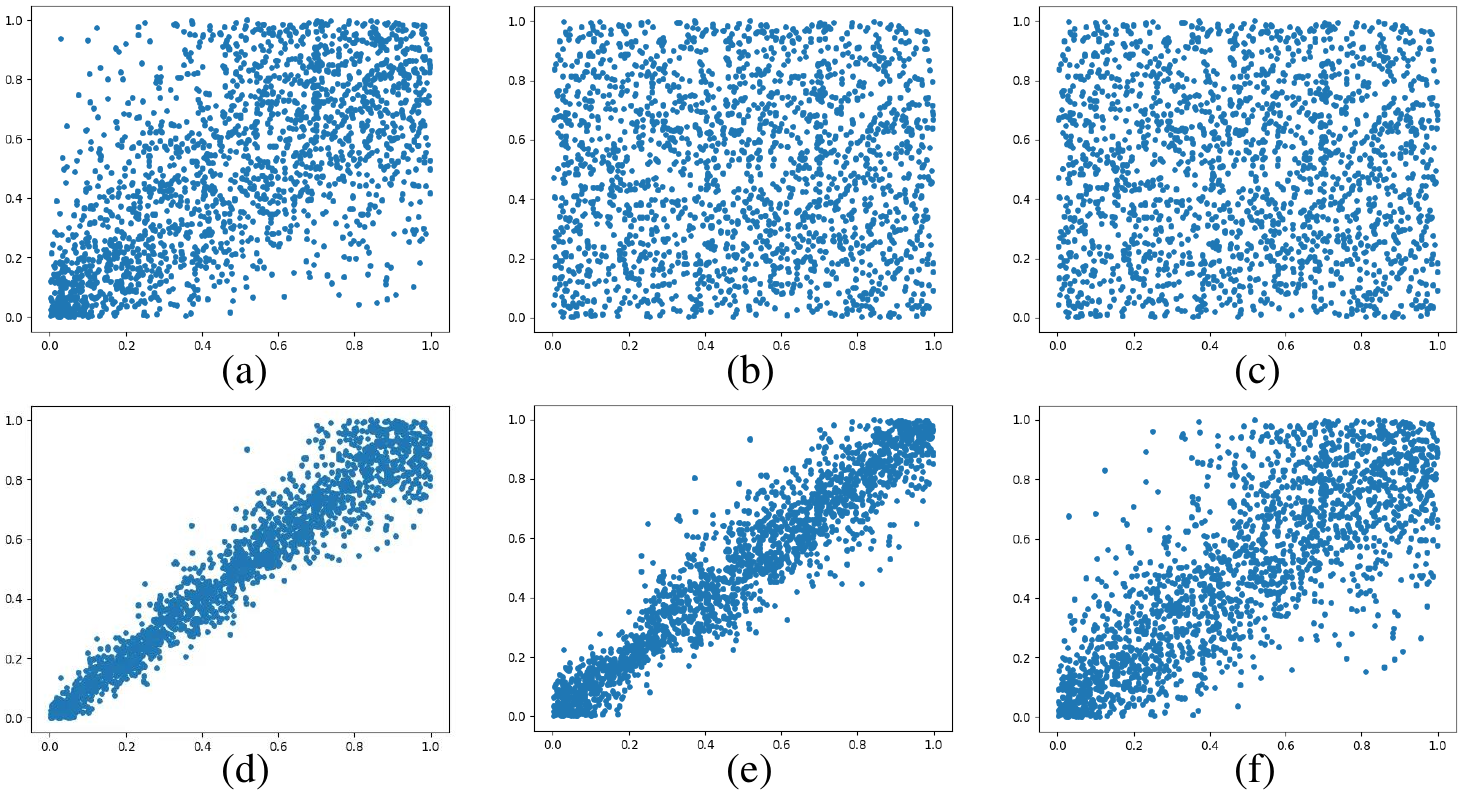}
    \caption{Copulas learned by HACSurv from Framingham and MIMIC-III dataset. (a), (b) and (c) are the copulas between Risk~1 and Risk~2, Risk~1 and censoring, and Risk~2 and censoring on the Framingham dataset. (d), (e) and (f) are the copulas between Risk~1 and Risk~3, Risk~3 and Risk~5, and Risk~2 and Risk~4 on MIMIC-III, respectively.}
    \label{fig:3}
\end{figure}

\begin{table*}[ht]
\centering 
\caption{Overall performance comparison on four datasets. We report the mean of the metrics for all risks. The results for each specific risk are provided in the supplementary materials. }
\label{tab:2}
\footnotesize
\setlength{\tabcolsep}{6pt} % Adjust column spacing
\begin{tabular}{llcccc}
\toprule
\textbf{Metric} & \textbf{Model} & \textbf{Synthetic} & \textbf{Framingham} & \textbf{SEER} & \textbf{MIMIC-III} \\
\midrule
\multirow{7}{*}{\(C^{td}\)-index}
& DeepHit       & $0.611 \pm 0.008$ & $0.699 \pm 0.019$ & $0.791 \pm 0.000$ & $0.765 \pm 0.015$ \\
& DSM           & $0.603 \pm 0.009$ & $0.734 \pm 0.016$ & $0.781 \pm 0.006$ & $0.752 \pm 0.018$ \\
& DeSurv        & $0.617 \pm 0.007$ & $0.674 \pm 0.022$ & $0.787 \pm 0.005$ & $0.755 \pm 0.021$ \\
& NeuralFG      & $0.549 \pm 0.019$ & $0.746 \pm 0.010$ & $0.779 \pm 0.004$ & $0.760 \pm 0.026$ \\
& HACSurv (I)   & $0.635 \pm 0.005$ & $\textbf{0.748} \pm 0.009$ & $0.797 \pm 0.003$ & $0.764 \pm 0.018$ \\
& HACSurv (S)   & $0.636 \pm 0.006$ & $0.746 \pm 0.013$ & $0.797 \pm 0.003$ & $0.760 \pm 0.016$ \\
& HACSurv       & $\textbf{0.643} \pm 0.007$ & $0.747 \pm 0.012$ & $\textbf{0.798} \pm 0.002$ & $\textbf{0.768} \pm 0.018$ \\
\midrule
\multirow{7}{*}{IBS}
& DeepHit       & $0.151 \pm 0.003$ & $0.089 \pm 0.003$ & $0.074 \pm 0.001$ & $0.264 \pm 0.083$ \\
& DSM           & $0.086 \pm 0.003$ & $0.088 \pm 0.004$ & $0.069 \pm 0.000$ & $0.226 \pm 0.080$ \\
& DeSurv        & $0.143 \pm 0.002$ & $0.104 \pm 0.004$ & $0.068 \pm 0.001$ & $0.268 \pm 0.085$ \\
& NeuralFG      & $0.173 \pm 0.016$ & $0.091 \pm 0.003$ & $0.068 \pm 0.001$ & $0.276 \pm 0.097$ \\
& HACSurv (I)   & $\textbf{0.077} \pm 0.002$ & $0.092 \pm 0.005$ & $\textbf{0.067} \pm 0.001$ & $0.169 \pm 0.049$ \\
& HACSurv (S)   & $0.078 \pm 0.004$ & $0.091 \pm 0.006$ & $\textbf{0.067} \pm 0.001$ & $\textbf{0.131} \pm 0.021$ \\
& HACSurv       & $0.100 \pm 0.008$ & $\textbf{0.085} \pm 0.003$ & $\textbf{0.067} \pm 0.001$ & $0.163 \pm 0.068$ \\
\bottomrule
\end{tabular}
\end{table*}

\paragraph{Qualitative Results}
As shown in Figure~\ref{fig:2}, the dependency structure learned by HACSurv from the synthetic dataset is almost identical to the true HAC. The results demonstrate that our model can learn complex dependency structures from partially observed competing risks survival data. The results from Figure~\ref{fig:3} on the Framingham dataset show that HACSurv found a positive dependency between Risk~1 and Risk~2, while both risks are independent of censoring. 
% Furthermore, by setting the copula between the two risks as an inner copula and the copula between each risk and censoring as an outer copula, HACSurv is capable of modeling such asymmetric dependencies in the Framingham dataset.

We present some of the dependency structures discovered in the MIMIC-III. Our model found strong positive dependencies between sepsis (Risk~1) and acute respiratory failure (Risk~3), as well as between acute respiratory failure (Risk~3) and pneumonia (Risk~5), and found a moderate positive dependency between cerebral hemorrhage (Risk~2) and subendocardial acute myocardial infarction (Risk~4). This is consistent with medical knowledge, demonstrating that HACSurv may uncover the interactions between diseases from partially observed data. 
% According to the dependencies between risks and censoring found in the first stage, in the second stage, HACSurv groups Risk1, 3, and 5 into one group, and Risk2 and 4 into another group, and models the dependency between the two groups and censoring using an outer copula. 

\begin{table} 
\small 
\centering
\caption{Performance comparison of survival models on synthetic dataset (Survival-$L_1$). In all tables, HACSurv (I) and HACSurv (S) respectively correspond to HACSurv with an independent copula and a single symmetry copula.}
\label{tab:1}
\setlength\tabcolsep{4pt} % Default value: 6pt
\renewcommand{\arraystretch}{1.1} % Default value: 1
\resizebox{1\columnwidth}{!}{%
\begin{tabular}{lccc}
\toprule
Method & Risk~1  & Risk~2  & Risk~3  \\
\midrule
DeepHit         & $0.375 \pm \scriptstyle{0.015}$ & $0.382 \pm \scriptstyle{0.015}$ & $0.152 \pm \scriptstyle{0.012}$ \\
DSM      & $0.243 \pm \scriptstyle{0.008}$ & $0.289 \pm \scriptstyle{0.005}$ & $0.067 \pm \scriptstyle{0.005}$ \\
DeSurv          & $0.362 \pm \scriptstyle{0.013}$ & $0.371 \pm \scriptstyle{0.012}$ & $0.133 \pm \scriptstyle{0.007}$ \\
NeuralFG        & $0.354 \pm \scriptstyle{0.012}$ & $0.359 \pm \scriptstyle{0.012}$ & $0.157 \pm \scriptstyle{0.004}$ \\
HACSurv (I) & $0.204 \pm \scriptstyle{0.006}$ & $0.230 \pm \scriptstyle{0.015}$ & $0.067 \pm \scriptstyle{0.001}$ \\
HACSurv (S)   & $0.039 \pm \scriptstyle{0.005}$ & $0.096 \pm \scriptstyle{0.008}$ & $0.013 \pm \scriptstyle{0.000}$ \\
HACSurv         & $\textbf{0.023} \pm \scriptstyle{0.004}$ & $\textbf{0.012} \pm \scriptstyle{0.002}$ & $\textbf{0.008} \pm \scriptstyle{0.002}$ \\
\bottomrule
\end{tabular}
}
\end{table}

% \begin{table}
% \centering
% \caption{Results of HACSurv Using Misspecified Copula, denoted as HACSurv (M). Only the \(C^{td}\)-index results are reported.}
% \label{tab:missspecified}
% \footnotesize
% \setlength{\tabcolsep}{8pt} % Adjust column spacing if needed
% \begin{tabular}{lcc}
% \toprule
% \textbf{Model} & \textbf{Framingham} & \textbf{MIMIC-III} \\
% \midrule
% HACSurv        & $0.747 \pm 0.012$   & $0.768 \pm 0.018$   \\
% HACSurv (I)    & $0.748 \pm 0.009$   & $0.764 \pm 0.018$   \\
% HACSurv (M)    & $0.744 \pm 0.012$   & $0.717 \pm 0.024$   \\
% \bottomrule
% \end{tabular}
% \end{table}

\begin{table}
\centering
\caption{Comparison of training time (per 100 epochs) and GPU memory usage on SEER dataset}
\resizebox{1\columnwidth}{!}{%
\begin{tabular}{lcccc}
\toprule
Method & Dim & Batch Size & Time (s) & Memory (MB)\\
\midrule
HACSurv & 2D & 20000 & 19.5 & 2890 \\
DCSurvival & 2D & 20000 & 255.36 & 21700 \\
HACSurv & 3D & 10000 & 38.05 & 3914 \\
DCSurvival & 3D & 10000 & 450.77 & 22356 \\
\bottomrule
\end{tabular}
\label{tab:3}
}
\end{table}

\paragraph{Quantitative Results} 
%%从这里开始写

The results in Table~\ref{tab:1} show that after identifying the dependency structure among events and censoring, HACSurv significantly reduces the estimation bias of the marginal survival distributions. Notably, even though only a single copula is used to capture the dependency, HACSurv (Symmetry) estimates the marginal survival distributions much better than HACSurv (Independent). DSM achieves better predictions than other existing methods due to its use of a parametric structure (Weibull) consistent with the true marginal distributions. However, since it trains and predicts based on CIF only and does not model the dependency structure, it is less effective than HACSurv.
\begin{figure*}[ht]
    \centering
    \includegraphics[width=0.70\linewidth]{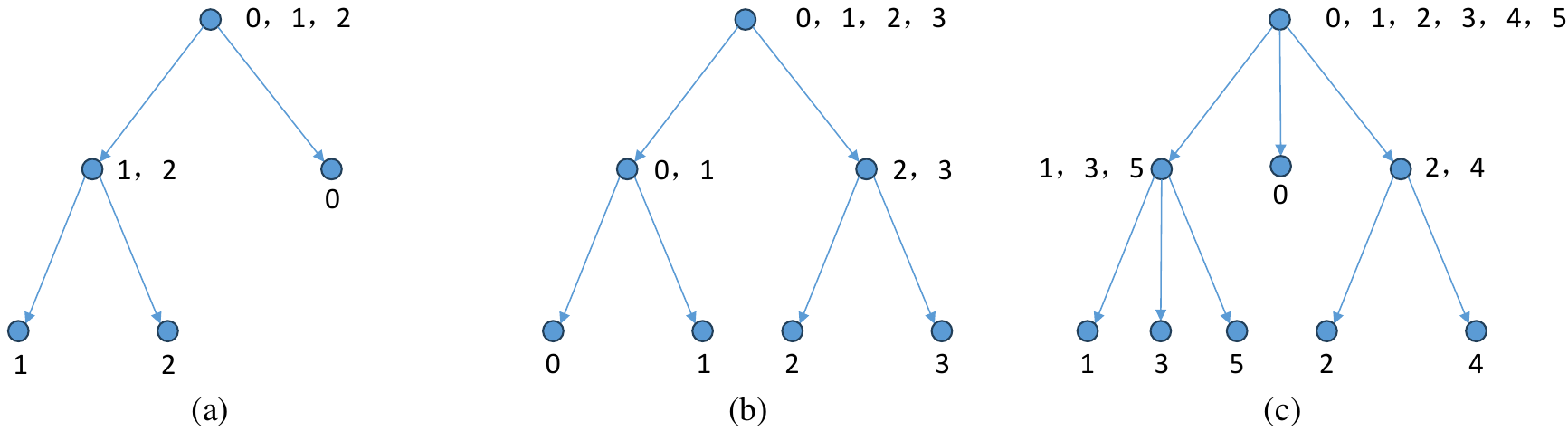}
    \caption{The HAC hierarchy determined using the HACSurv framework: (a) Framingham and SEER datasets, (b) synthetic dataset, and (c) MIMIC-III dataset.}
    \label{fig:HACstructure}
\end{figure*}

According to Table~\ref{tab:2}, HACSurv significantly outperforms existing methods in terms of the \( C^{td} \)-index metric. For IBS, HACSurv also achieves results superior to existing state-of-the-art methods. However, it sometimes performs worse than HACSurv (Independent).
% According to Table~\ref{tab:2}, HACSurv significantly outperforms existing methods in terms of the \( C^{td} \)-index metric. It is worth mentioning that although HACSurv achieves near-perfect survival marginal predictions on the synthetic dataset, its IBS is not as good as its version based on an independent copula. This further confirms that IBS is not a strictly proper scoring rule. Nevertheless, in terms of IBS, HACSurv still achieves results superior to existing state-of-the-art methods.

Table~\ref{tab:3} shows the computational efficiency comparison between HACSurv and DCSurvival in the case of a single copula. The training time of HACSurv is only about 8\% of that of DCSurvival, with 17\% of the GPU memory usage. Thus, in the single-risk setting, HACSurv also represents an improvement over DCSurvival.

% , due to the higher computational efficiency of Gen-AC~\cite{ng2021generative} than AC-Net~\cite{ling2020deepAC}.

\paragraph{IBS is Not a Proper Scoring Rule}
It is worth mentioning that although HACSurv achieves near-perfect survival marginal predictions on the synthetic dataset, its IBS is not as good as its version based on an independent copula. Moreover, on real-world datasets, HACSurv does not achieve the best results.

When dependent censoring exists, IBS may not be a reliable metric because its calculation uses Inverse Probability of Censoring Weighting (IPCW)~\citep{kvamme2023brier}, which relies on the independent censoring assumption. In our experiments, for the synthetic and MIMIC-III data sets, there are complex dependencies between censoring and competing risks, making IBS potentially unreliable. However, for the Framingham and SEER datasets, we observed that censoring is almost independent of competing risks, aligning with the independent censoring assumption. In this case, HACSurv achieves relatively better performance.

\paragraph{Determining the Hierarchical Structure of HAC}\label{sec:HACstructure}
The determination of the HAC’s hierarchical structure should be flexible. Sometimes, medical knowledge can also serve as a reference. As shown in Figure~\ref{fig:HACstructure}, we present the hierarchical structure in our experiments as an example for practitioners.

In our experiments on the Framingham and SEER datasets, we found dependencies between the two competing risks, while the correlations between risks and censoring were nearly independent. Therefore, we selected the copula between competing risks as inner. For the MIMIC-III dataset, we observed strong pairwise correlations between risk~1, risk~2, and risk~5, which are related to the respiratory system. Thus, we set up an inner copula to capture their dependencies. Furthermore, the copula between risk~2 and risk~4 also exhibited a strong correlation, so we chose it as the second inner copula. We used the weakest copula between risks and censoring as the outer copula.

\begin{table}
\centering
\caption{Results of HACSurv using misspecified copula, denoted as HACSurv (M). Only the \(C^{td}\)-index results are reported.}
\label{tab:missspecified}
\footnotesize
\setlength{\tabcolsep}{8pt} % Adjust column spacing if needed
\begin{tabular}{lcc}
\toprule
\textbf{Model} & \textbf{Framingham} & \textbf{MIMIC-III} \\
\midrule
HACSurv        & $0.747 \pm 0.012$   & $0.768 \pm 0.018$   \\
HACSurv (I)    & $0.748 \pm 0.009$   & $0.764 \pm 0.018$   \\
HACSurv (M)    & $0.744 \pm 0.012$   & $0.717 \pm 0.024$   \\
\bottomrule
\end{tabular}
\end{table}

\paragraph{Impact of Misspecified Copula}
We caution that misspecifying the copula may lead to worse results. In Table~\ref{tab:missspecified}, for the Framingham experiment, we deliberately misspecify the copula in Figure~\ref{fig:3} (a) as the outer copula between competing events and censoring, while setting the inner copula between competing events to one with stronger dependency. For the MIMIC-III experiment, we use a copula with strong dependency, similar to the one in Figure~\ref{fig:3} (f), to model symmetric dependency. The results in Table~\ref{tab:missspecified} show that using an obviously incorrect copula to model dependencies leads to survival prediction performance that is much worse than either assuming independent risks or using the asymmetric dependencies identified by our method.

We provide detailed experimental settings, additional qualitative results, and an analysis of the differences in predictions using marginal distributions and CIF in the supplementary materials.

\section{CONCLUSION}
In this paper, we propose HACSurv, a survival analysis method capable of modeling the dependency structure among competing risks and censoring. Our method directly uses the survival likelihood for training, without relying on the independent censoring assumption or the independent competing risks assumption. We use hierarchical Archimedean copulas (HACs) to flexibly model asymmetric dependency structures. HACSurv first learns the structure and parameters of the HAC from partially observed survival data and then learns the marginal survival distributions. This paper also explores the prediction objective under the survival analysis setting, extending the use of survival marginal for prediction in single-risk scenarios to using conditional cause-specific CIF to better model the interactions among competing risks. Empirically, we demonstrate that HACSurv can closely approximate the dependency structures in survival data and significantly reduce survival estimation bias.

However, there are some limitations in our proposed HACSurv. Although HACs are capable of characterizing most dependencies in the real world, there are certain dependency structures that cannot be accurately represented by HACs. A possible future direction is to explore vine copulas that are more flexible than HAC.

\clearpage
\bibliographystyle{plainnat}

\bibliography{references}

\begin{thebibliography}{42}
\providecommand{\natexlab}[1]{#1}
\providecommand{\url}[1]{\texttt{#1}}
\expandafter\ifx\csname urlstyle\endcsname\relax
  \providecommand{\doi}[1]{doi: #1}\else
  \providecommand{\doi}{doi: \begingroup \urlstyle{rm}\Url}\fi

\bibitem[Antolini et~al.(2005)Antolini, Boracchi, and Biganzoli]{antolini2005cindex}
Laura Antolini, Patrizia Boracchi, and Elia Biganzoli.
\newblock A time-dependent discrimination index for survival data.
\newblock \emph{Statistics in medicine}, 24\penalty0 (24):\penalty0 3927--3944, 2005.

\bibitem[Bernstein(1929)]{bernstein1929fonctions}
Serge Bernstein.
\newblock Sur les fonctions absolument monotones.
\newblock \emph{Acta Mathematica}, 52\penalty0 (1):\penalty0 1--66, 1929.

\bibitem[Bosco~Sabuhoro et~al.(2006)Bosco~Sabuhoro, Larue, and Gervais]{bosco2006factorsfinance2}
Jean Bosco~Sabuhoro, Bruno Larue, and Yvan Gervais.
\newblock Factors determining the success or failure of canadian establishments on foreign markets: A survival analysis approach.
\newblock \emph{The International Trade Journal}, 20\penalty0 (1):\penalty0 33--73, 2006.

\bibitem[Caselli et~al.(2021)Caselli, Corbetta, Cucinelli, and Rossolini]{caselli2021survivalfinace1}
Stefano Caselli, Guido Corbetta, Doriana Cucinelli, and Monica Rossolini.
\newblock A survival analysis of public guaranteed loans: Does financial intermediary matter?
\newblock \emph{Journal of Financial Stability}, 54:\penalty0 100880, 2021.

\bibitem[Chilinski and Silva(2020)]{chilinski2020neural}
Pawel Chilinski and Ricardo Silva.
\newblock Neural likelihoods via cumulative distribution functions.
\newblock In \emph{Conference on Uncertainty in Artificial Intelligence}, pages 420--429. PMLR, 2020.

\bibitem[Cox(1972)]{cox1972COXPH}
David~R Cox.
\newblock Regression models and life-tables.
\newblock \emph{Journal of the Royal Statistical Society: Series B (Methodological)}, 34\penalty0 (2):\penalty0 187--202, 1972.

\bibitem[Danks and Yau(2022)]{danks2022DeSurv}
Dominic Danks and Christopher Yau.
\newblock Derivative-based neural modelling of cumulative distribution functions for survival analysis.
\newblock In \emph{International Conference on Artificial Intelligence and Statistics}, pages 7240--7256. PMLR, 2022.

\bibitem[Davis et~al.(2014)Davis, Li, Wai, Tyldesley, Simmons, Baliski, and McBride]{davis2014BreastSurvivor}
Margot~K Davis, Dongdong Li, Elaine Wai, Scott Tyldesley, Christine Simmons, Christopher Baliski, and Mary~L McBride.
\newblock Hospital-related cardiac morbidity among survivors of breast cancer: Long-term risks and predictors.
\newblock \emph{Journal of Cardiac Failure}, 20\penalty0 (8):\penalty0 S44--S45, 2014.

\bibitem[Emura and Chen(2018)]{emura2018analysisofSurvival}
Takeshi Emura and Yi-Hau Chen.
\newblock \emph{Analysis of survival data with dependent censoring: copula-based approaches}, volume 450.
\newblock Springer, 2018.

\bibitem[Fine and Gray(1999)]{fine1999proportional}
Jason~P Fine and Robert~J Gray.
\newblock A proportional hazards model for the subdistribution of a competing risk.
\newblock \emph{Journal of the American statistical association}, 94\penalty0 (446):\penalty0 496--509, 1999.

\bibitem[Friedman et~al.(2015)Friedman, Furberg, DeMets, Reboussin, and Granger]{friedman2015medicine1}
Lawrence~M Friedman, Curt~D Furberg, David~L DeMets, David~M Reboussin, and Christopher~B Granger.
\newblock \emph{Fundamentals of clinical trials}.
\newblock Springer, 2015.

\bibitem[Gharari et~al.(2023)Gharari, Cooper, Greiner, and Krishnan]{gharari2023copula}
Ali Hossein~Foomani Gharari, Michael Cooper, Russell Greiner, and Rahul~G Krishnan.
\newblock Copula-based deep survival models for dependent censoring.
\newblock In \emph{Uncertainty in Artificial Intelligence}, pages 669--680. PMLR, 2023.

\bibitem[G{\'o}recki et~al.(2017)G{\'o}recki, Hofert, and Hole{\v{n}}a]{gorecki2017structure}
Jan G{\'o}recki, Marius Hofert, and Martin Hole{\v{n}}a.
\newblock On structure, family and parameter estimation of hierarchical archimedean copulas.
\newblock \emph{Journal of Statistical Computation and Simulation}, 87\penalty0 (17):\penalty0 3261--3324, 2017.

\bibitem[Graf et~al.(1999)Graf, Schmoor, Sauerbrei, and Schumacher]{graf1999IBS}
Erika Graf, Claudia Schmoor, Willi Sauerbrei, and Martin Schumacher.
\newblock Assessment and comparison of prognostic classification schemes for survival data.
\newblock \emph{Statistics in medicine}, 18\penalty0 (17-18):\penalty0 2529--2545, 1999.

\bibitem[Hering et~al.(2010)Hering, Hofert, Mai, and Scherer]{hering2010constructing}
Christian Hering, Marius Hofert, Jan-Frederik Mai, and Matthias Scherer.
\newblock Constructing hierarchical archimedean copulas with l{\'e}vy subordinators.
\newblock \emph{Journal of Multivariate Analysis}, 101\penalty0 (6):\penalty0 1428--1433, 2010.

\bibitem[Howlader et~al.(2010)Howlader, Ries, Mariotto, Reichman, Ruhl, and Cronin]{howlader2010improved}
Nadia Howlader, Lynn~AG Ries, Angela~B Mariotto, Marsha~E Reichman, Jennifer Ruhl, and Kathleen~A Cronin.
\newblock Improved estimates of cancer-specific survival rates from population-based data.
\newblock \emph{JNCI: Journal of the National Cancer Institute}, 102\penalty0 (20):\penalty0 1584--1598, 2010.

\bibitem[Jeanselme et~al.(2023)Jeanselme, Yoon, Tom, and Barrett]{jeanselme2023neuralFG}
Vincent Jeanselme, Chang~Ho Yoon, Brian Tom, and Jessica Barrett.
\newblock Neural fine-gray: Monotonic neural networks for competing risks.
\newblock In \emph{Conference on Health, Inference, and Learning}, pages 379--392. PMLR, 2023.

\bibitem[Joe(1997)]{joe1997multivariate}
Harry Joe.
\newblock \emph{Multivariate models and multivariate dependence concepts}.
\newblock CRC press, 1997.

\bibitem[Kannel and McGee(1979)]{kannel1979Framingham}
William~B Kannel and Daniel~L McGee.
\newblock Diabetes and cardiovascular disease: the framingham study.
\newblock \emph{Jama}, 241\penalty0 (19):\penalty0 2035--2038, 1979.

\bibitem[Ken-Iti(1999)]{ken1999levy}
Sato Ken-Iti.
\newblock \emph{L{\'e}vy processes and infinitely divisible distributions}, volume~68.
\newblock Cambridge university press, 1999.

\bibitem[Kvamme et~al.(2019)Kvamme, {{\O}}rnulf Borgan, and Scheel]{Kvamme2019}
H{{\aa}}vard Kvamme, {{\O}}rnulf Borgan, and Ida Scheel.
\newblock Time-to-event prediction with neural networks and cox regression.
\newblock \emph{Journal of Machine Learning Research}, 20\penalty0 (129):\penalty0 1--30, 2019.
\newblock URL \url{http://jmlr.org/papers/v20/18-424.html}.

\bibitem[Kvamme and Borgan(2023)]{kvamme2023brier}
H{\aa}vard Kvamme and {\O}rnulf Borgan.
\newblock The brier score under administrative censoring: Problems and a solution.
\newblock \emph{Journal of Machine Learning Research}, 24\penalty0 (2):\penalty0 1--26, 2023.

\bibitem[Lee et~al.(2018)Lee, Zame, Yoon, and Van Der~Schaar]{lee2018deephit}
Changhee Lee, William Zame, Jinsung Yoon, and Mihaela Van Der~Schaar.
\newblock Deephit: A deep learning approach to survival analysis with competing risks.
\newblock In \emph{Proceedings of the AAAI conference on artificial intelligence}, volume~32, 2018.

\bibitem[Li et~al.(2023)Li, Hu, and Wang]{li2023evaluating}
Dongdong Li, X~Joan Hu, and Rui Wang.
\newblock Evaluating association between two event times with observations subject to informative censoring.
\newblock \emph{Journal of the American Statistical Association}, 118\penalty0 (542):\penalty0 1282--1294, 2023.

\bibitem[Li and Lu(2019)]{li2019modelingCause}
Hong Li and Yang Lu.
\newblock Modeling cause-of-death mortality using hierarchical archimedean copula.
\newblock \emph{Scandinavian Actuarial Journal}, 2019\penalty0 (3):\penalty0 247--272, 2019.

\bibitem[Ling et~al.(2020)Ling, Fang, and Kolter]{ling2020deepAC}
Chun~Kai Ling, Fei Fang, and J~Zico Kolter.
\newblock Deep archimedean copulas.
\newblock \emph{Advances in Neural Information Processing Systems}, 33:\penalty0 1535--1545, 2020.

\bibitem[Lo et~al.(2020)Lo, Mammen, and Wilke]{lo2020nested}
Simon~MS Lo, Enno Mammen, and Ralf~A Wilke.
\newblock A nested copula duration model for competing risks with multiple spells.
\newblock \emph{Computational Statistics \& Data Analysis}, 150:\penalty0 106986, 2020.

\bibitem[Loshchilov and Hutter(2018)]{loshchilov2018AdamW}
Ilya Loshchilov and Frank Hutter.
\newblock Decoupled weight decay regularization.
\newblock In \emph{International Conference on Learning Representations}, 2018.

\bibitem[McNeil(2008)]{mcneil2008sampling}
Alexander~J McNeil.
\newblock Sampling nested archimedean copulas.
\newblock \emph{Journal of Statistical Computation and Simulation}, 78\penalty0 (6):\penalty0 567--581, 2008.

\bibitem[Nagpal et~al.(2021)Nagpal, Li, and Dubrawski]{nagpal2021deep}
Chirag Nagpal, Xinyu Li, and Artur Dubrawski.
\newblock Deep survival machines: Fully parametric survival regression and representation learning for censored data with competing risks.
\newblock \emph{IEEE Journal of Biomedical and Health Informatics}, 25\penalty0 (8):\penalty0 3163--3175, 2021.

\bibitem[Ng et~al.(2021)Ng, Hasan, Elkhalil, and Tarokh]{ng2021generative}
Yuting Ng, Ali Hasan, Khalil Elkhalil, and Vahid Tarokh.
\newblock Generative archimedean copulas.
\newblock In \emph{Uncertainty in Artificial Intelligence}, pages 643--653. PMLR, 2021.

\bibitem[Okhrin et~al.(2017)Okhrin, Ristig, and Xu]{okhrin2017CopulaeinHighDimensions}
Ostap Okhrin, Alexander Ristig, and Ya-Fei Xu.
\newblock Copulae in high dimensions: an introduction.
\newblock \emph{Applied quantitative finance}, pages 247--277, 2017.

\bibitem[Paszke et~al.(2017)Paszke, Gross, Chintala, Chanan, Yang, DeVito, Lin, Desmaison, Antiga, and Lerer]{paszke2017automatic}
Adam Paszke, Sam Gross, Soumith Chintala, Gregory Chanan, Edward Yang, Zachary DeVito, Zeming Lin, Alban Desmaison, Luca Antiga, and Adam Lerer.
\newblock Automatic differentiation in pytorch.
\newblock 2017.

\bibitem[Pe{\~n}a and Hollander(2004)]{pena2004modelsreliability}
Edsel~A Pe{\~n}a and Myles Hollander.
\newblock Models for recurrent events in reliability and survival analysis.
\newblock \emph{Mathematical reliability: An expository perspective}, pages 105--123, 2004.

\bibitem[Rindt et~al.(2022)Rindt, Hu, Steinsaltz, and Sejdinovic]{rindt2022survival}
David Rindt, Robert Hu, David Steinsaltz, and Dino Sejdinovic.
\newblock Survival regression with proper scoring rules and monotonic neural networks.
\newblock In \emph{International Conference on Artificial Intelligence and Statistics}, pages 1190--1205. PMLR, 2022.

\bibitem[Tang et~al.(2021)Tang, Davarmanesh, Song, Koutra, Sjoding, and Wiens]{tang2021mimic}
Shengpu Tang, Parmida Davarmanesh, Yanmeng Song, Danai Koutra, Michael Sjoding, and Jenna Wiens.
\newblock Mimic-iii and eicu-crd: Feature representation by fiddle preprocessing.
\newblock \emph{Shock (4h)}, 19\penalty0 (98):\penalty0 4--522, 2021.

\bibitem[Tankov(2003)]{tankov2003financial}
Peter Tankov.
\newblock \emph{Financial modelling with jump processes}.
\newblock Chapman and Hall/CRC, 2003.

\bibitem[Tsiatis(1975)]{tsiatis1975nonidentifiability}
Anastasios Tsiatis.
\newblock A nonidentifiability aspect of the problem of competing risks.
\newblock \emph{Proceedings of the National Academy of Sciences}, 72\penalty0 (1):\penalty0 20--22, 1975.

\bibitem[Wang and Sun(2022)]{wang2022survtrace}
Zifeng Wang and Jimeng Sun.
\newblock Survtrace: Transformers for survival analysis with competing events.
\newblock In \emph{Proceedings of the 13th ACM international conference on bioinformatics, computational biology and health informatics}, pages 1--9, 2022.

\bibitem[Widder(2015)]{widder2015laplaceTrans}
David~Vernon Widder.
\newblock Laplace transform (pms-6).
\newblock 2015.

\bibitem[Yeh et~al.(2016)Yeh, Secemsky, Kereiakes, Normand, Gershlick, Cohen, Spertus, Steg, Cutlip, Rinaldi, et~al.]{yeh2016developmentmedicine2}
Robert~W Yeh, Eric~A Secemsky, Dean~J Kereiakes, Sharon-Lise~T Normand, Anthony~H Gershlick, David~J Cohen, John~A Spertus, Philippe~Gabriel Steg, Donald~E Cutlip, Michael~J Rinaldi, et~al.
\newblock Development and validation of a prediction rule for benefit and harm of dual antiplatelet therapy beyond 1 year after percutaneous coronary intervention.
\newblock \emph{Jama}, 315\penalty0 (16):\penalty0 1735--1749, 2016.

\bibitem[Zhang et~al.(2024)Zhang, Ling, and Zhang]{zhang2024DCSurvival}
Weijia Zhang, Chun~Kai Ling, and Xuanhui Zhang.
\newblock Deep copula-based survival analysis for dependent censoring with identifiability guarantees.
\newblock In \emph{Proceedings of the AAAI Conference on Artificial Intelligence}, volume~38, pages 20613--20621, 2024.

\end{thebibliography}
\clearpage
\section*{Checklist}

 \begin{enumerate}

 \item For all models and algorithms presented, check if you include:
 \begin{enumerate}
   \item A clear description of the mathematical setting, assumptions, algorithm, and/or model. [Yes]
   \item An analysis of the properties and complexity (time, space, sample size) of any algorithm. [Yes]
   \item (Optional) Anonymized source code, with specification of all dependencies, including external libraries. [Yes]
 \end{enumerate}

 \item For any theoretical claim, check if you include:
 \begin{enumerate}
   \item Statements of the full set of assumptions of all theoretical results. [Yes]
   \item Complete proofs of all theoretical results. [Yes]
   \item Clear explanations of any assumptions. [Yes]     
 \end{enumerate}

 \item For all figures and tables that present empirical results, check if you include:
 \begin{enumerate}
   \item The code, data, and instructions needed to reproduce the main experimental results (either in the supplemental material or as a URL). [Yes]
   \item All the training details (e.g., data splits, hyperparameters, how they were chosen). [Yes/No/Not Applicable]
         \item A clear definition of the specific measure or statistics and error bars (e.g., with respect to the random seed after running experiments multiple times). [Yes]
         \item A description of the computing infrastructure used. (e.g., type of GPUs, internal cluster, or cloud provider). [Yes]
 \end{enumerate}

 \item If you are using existing assets (e.g., code, data, models) or curating/releasing new assets, check if you include:
 \begin{enumerate}
   \item Citations of the creator If your work uses existing assets. [Yes]
   \item The license information of the assets, if applicable. [Yes]
   \item New assets either in the supplemental material or as a URL, if applicable. [Yes]
   \item Information about consent from data providers/curators. [Yes]
   \item Discussion of sensible content if applicable, e.g., personally identifiable information or offensive content. [Not Applicable]
 \end{enumerate}

 \item If you used crowdsourcing or conducted research with human subjects, check if you include:
 \begin{enumerate}
   \item The full text of instructions given to participants and screenshots. [Not Applicable]
   \item Descriptions of potential participant risks, with links to Institutional Review Board (IRB) approvals if applicable. [Not Applicable]
   \item The estimated hourly wage paid to participants and the total amount spent on participant compensation. [Not Applicable]
 \end{enumerate}

 \end{enumerate}

% \subsubsection*{References}

% References follow the acknowledgements.  Use an unnumbered third level
% heading for the references section.  Please use the same font
% size for references as for the body of the paper---remember that
% references do not count against your page length total.

% \begin{thebibliography}{}
% \setlength{\itemindent}{-\leftmargin}
% \makeatletter\renewcommand{\@biblabel}[1]{}\makeatother
% \bibitem{} J.~Alspector, B.~Gupta, and R.~B.~Allen (1989).
%     \newblock Performance of a stochastic learning microchip.
%     \newblock In D. S. Touretzky (ed.),
%     \textit{Advances in Neural Information Processing Systems 1}, 748--760.
%     San Mateo, Calif.: Morgan Kaufmann.

% \bibitem{} F.~Rosenblatt (1962).
%     \newblock \textit{Principles of Neurodynamics.}
%     \newblock Washington, D.C.: Spartan Books.

% \bibitem{} G.~Tesauro (1989).
%     \newblock Neurogammon wins computer Olympiad.
%     \newblock \textit{Neural Computation} \textbf{1}(3):321--323.
% \end{thebibliography}

%%%%%%%%%%%%%%%%%%%%%%%%%%%%%%%%%%%%%%%%%%%%%%%%%%%%%%%%%%%%
\clearpage

\appendix
\onecolumn
\aistatstitle{
Supplementary Materials}
% \onecolumn 
% \section{Appendix}

% \section{DERIVATION OF SURVIVAL LIKELIHOOD}
\section{DERIVATION OF SURVIVAL LIKELIHOOD}
We omit covariates $\boldsymbol{x}$ for brevity. We start with the likelihood function:
\begin{equation}
\mathcal{L} = \prod_{k=0}^{K} \left[ \Pr(T_k = t, \{T_i > t\}_{i \neq k}) \right]^{\mathbb{1}_{\{e=k\}}}
\end{equation}
\hfill (Initial likelihood expression)

Applying the definition of the survival function and the relationship between probability density and survival functions, we rewrite the probability as:
\begin{equation}
\mathcal{L} = \prod_{k=0}^{K} \left[ -\frac{d}{dy} \Pr(T_k > y, \{T_i > t\}_{i \neq k}) \Bigg|_{y = t} \right]^{\mathbb{1}_{\{e=k\}}}
\end{equation}
\hfill (Definition of Survival Function)

Using Sklar's Theorem to express the joint survival function in terms of a copula function \( C(u_0, \dots, u_K) \), where \( u_k = S_{T_k}(y) \) and \( u_i = S_{T_i}(t) \) for \( i \neq k \), we have:
\begin{equation}
\mathcal{L} = \prod_{k=0}^{K} \left[ -\frac{d}{dy} C(u_0, \dots, u_K) \Bigg|_{\substack{u_k = S_{T_k}(y) \\ y = t \\ u_i = S_{T_i}(t),\, i \neq k}} \right]^{\mathbb{1}_{\{e=k\}}}
\end{equation}
\hfill (Sklar's Theorem)

Applying the Chain Rule of differentiation, we obtain:
\begin{equation}
\mathcal{L} = \prod_{k=0}^{K} \left[ f_{T_k}(t) \cdot \frac{\partial}{\partial u_k} C(u_0, \dots, u_K) \Bigg|_{u_i = S_{T_i}(t)} \right]^{\mathbb{1}_{\{e=k\}}}
\end{equation}
\hfill (Chain Rule of derivative)

\section{DERIVATION OF THE CONDITIONAL CIF}

We omit covariates $\boldsymbol{x}^*$ and derive the cause-specific conditional cumulative incidence function (CIF):
\begin{equation}
F_{k^*}(t^*) = \Pr(T_{k^*} < t^* \mid \{T_i > t^*\}_{i \neq k^*}).
\end{equation}
\hfill (Initial definition)

By definition of conditional probability, it can be rewritten as:
\begin{equation}
F_{k^*}(t^*) = \frac{\Pr(T_{k^*} < t^*, \{T_i > t^*\}_{i \neq k^*})}{\Pr(\{T_i > t^*\}_{i \neq k^*})}.
\end{equation}
\hfill (Conditional probability)

Using Sklar's theorem, express the joint survival function in terms of copula:
\begin{equation}
\Pr(\{T_i > t^*\}_{i = 0, \dots, K}) = C(u_0, \dots, u_K), \quad \Pr(\{T_i > t^*\}_{i \neq k^*}) = C(\{u_i\}_{i \neq k^*}).
\end{equation}
\hfill (Sklar's theorem)
\clearpage
The joint probability \(\Pr(T_{k^*} < t^*, \{T_i > t^*\}_{i \neq k^*})\) is given by:
\begin{equation}
\Pr(T_{k^*} < t^*, \{T_i > t^*\}_{i \neq k^*}) = C(\{u_i\}_{i \neq k^*}) - C(u_0, \dots, u_K).
\end{equation}
% \hfill (Difference of joint survival probabilities)

Substituting into the conditional probability formula and simplifying yields:
\begin{equation}
F_{k^*}(t^*) = 1 - \frac{C(\{u_i\}_{i = 0, \dots, K})}{C(\{u_i\}_{i \neq k^*})}.
\end{equation}
where each \(u_i = S_{T_i \mid X}(t^* \mid \boldsymbol{x}^*)\) for all \(i = 0, \dots, K\).
\hfill (Final simplified result)

\section{USING LÉVY SUBORDINATORS TO CONSTRUCT INNER GENERATORS OF HIERARCHICAL ARCHIMEDEAN COPULAS}

The criterion that the derivative \(\left(\varphi_0^{-1} \circ \varphi_j\right)^{\prime}\) are completely monotone, is addressed in~\cite{hering2010constructing}, using Lévy Subordinators, i.e. non-decreasing Lévy processes such as the compound Poisson process, by recognizing that the Laplace transform of Lévy subordinators at a given ‘time’ \(t \geq 0\) have the form \(e^{-t \psi_j}\), where the Laplace exponent \(\psi_j\) has completely monotone derivative. For more background on Lévy processes, we refer the reader to the books~\cite{ken1999levy,tankov2003financial}. Here we restate the method proposed by~\cite{hering2010constructing} and utilized in~\cite{ng2021generative}.

For a given outer generator  \(\varphi_0\), a compatible inner generator \(\varphi_j\) can be constructed by composing the outer generator with the 
Laplace exponent \(\psi_j\) of a Lévy subordinator:
\begin{equation}
\varphi_j(x) = (\varphi_0 \circ \psi_j)(x)
\end{equation}

where the Laplace exponent \(\psi_j : [0, \infty) \to [0, \infty)\) of a Lévy subordinator has a convenient representation with drift \(\mu_j \geq 0\) and Lévy measure \(\nu_j\) on \((0, \infty)\) due to the Lévy-Khintchine 
theorem~\cite{ken1999levy}:

\begin{equation}
\psi_j(x)=\mu_j x+\int_0^{\infty}\left(1-e^{-x s}\right) \nu_j(d s)
 \end{equation}

A popular Lévy subordinator is the compound Poisson process with drift $\mu_j \geq 0$, jump intensity $\beta_j>0$, and jump size distribution determined by its Laplace transform $\varphi_{M_j}$. In this case, the Laplace exponent has the following expression:

\begin{equation}
\begin{aligned}
\psi_j(x) &= \mu_j x + \beta_j \left(1 - \varphi_{M_j}(x)\right) \\
          &= \mu_j x + \beta_j \left(1 - \int_0^{\infty} e^{-x s} \, dF_{M_j}(s)\right)
\end{aligned}
\end{equation}

where $M_j>0$ is a positive random variable with Laplace transform $\varphi_{M_j}$ characterizing the jump sizes of the compound Poisson process.
In addition, we choose $\mu_j>0$ to satisfy the condition $\varphi_j(\infty)=\left(\varphi_0 \circ \psi_j\right)(\infty)=0$ such that $\varphi_j$ is a valid generator of an Archimedean copula. For the sampling algorithm of HACs, we refer the reader to~\cite{hering2010constructing}.

\section{ADDITIONAL EXPERIMENTS}
\subsection{Comparison of Marginal Survival Function and CIF}
Table~\ref{tab:comparison_SF_CIF} presents the results of using marginal survival functions and CIF for predictions across four datasets. For the model based on the independent copula, there is no difference between using marginal survival functions and CIF for predictions. However, for the HAC-based model, whether marginal survival functions or CIF yield better performance varies across different datasets. We believe this may be due to the complexity of survival analysis tasks and differences in data collection across datasets. Therefore, we suggest that the final choice of whether to use the marginal survival function or CIF for prediction should be based on the results on the validation set.

\begin{table}[ht]
\centering 
\caption{Comparison of Marginal Survival Function and CIF on Four Datasets. SF corresponds to predictions using the marginal survival function. CIF corresponds to predictions using the cause-specific conditional CIF.}
\label{tab:comparison_SF_CIF}
\footnotesize
\setlength{\tabcolsep}{6pt} % Adjust column spacing
\begin{tabular}{llcccc}
\toprule
\textbf{Metric} & \textbf{Model} & \textbf{Synthetic} & \textbf{Framingham} & \textbf{SEER} & \textbf{MIMIC-III} \\
\midrule
\multirow{4}{*}{\(C^{td}\)-index}
& HACSurv + SF             & $0.5987 \pm 0.005$ & $0.7467 \pm 0.011$ & $0.7962 \pm 0.002$ & $\textbf{0.7682} \pm 0.018$ \\
& HACSurv + CIF        & $\textbf{0.6426} \pm 0.007$ & $0.7415 \pm 0.012$ & $\textbf{0.7975} \pm 0.002$ & $0.7506 \pm 0.017$ \\
& HACSurv (I) + SF          & $0.6350 \pm 0.005$ & $\textbf{0.7478} \pm 0.009$ & $0.7972 \pm 0.003$ & $0.7635 \pm 0.018$ \\
& HACSurv (I) + CIF    & $0.6350 \pm 0.005$ & $\textbf{0.7478} \pm 0.009$ & $0.7972 \pm 0.003$ & $0.7635 \pm 0.018$ \\
\midrule
\multirow{4}{*}{IBS}
& HACSurv + SF              & $0.1205 \pm 0.005$ & $\textbf{0.0850} \pm 0.003$ & $0.0669 \pm 0.000$ & $0.1629 \pm 0.068$ \\
& HACSurv + CIF        & $0.0999 \pm 0.008$ & $0.0930 \pm 0.003$ & $0.0670 \pm 0.000$ & $\textbf{0.1540} \pm 0.058$ \\
& HACSurv (I) + SF          & $\textbf{0.0773} \pm 0.002$ & $0.0922 \pm 0.004$ & $\textbf{0.0668} \pm 0.000$ & $0.1690 \pm 0.049$ \\
& HACSurv (I) + CIF    & $\textbf{0.0773} \pm 0.002$ & $0.0922 \pm 0.004$ & $\textbf{0.0668} \pm 0.000$ & $0.1690 \pm 0.049$ \\
\bottomrule
\end{tabular}
\end{table}

% \section{Conclusion}
\begin{figure}
    \centering
    \includegraphics[width=0.6\linewidth]{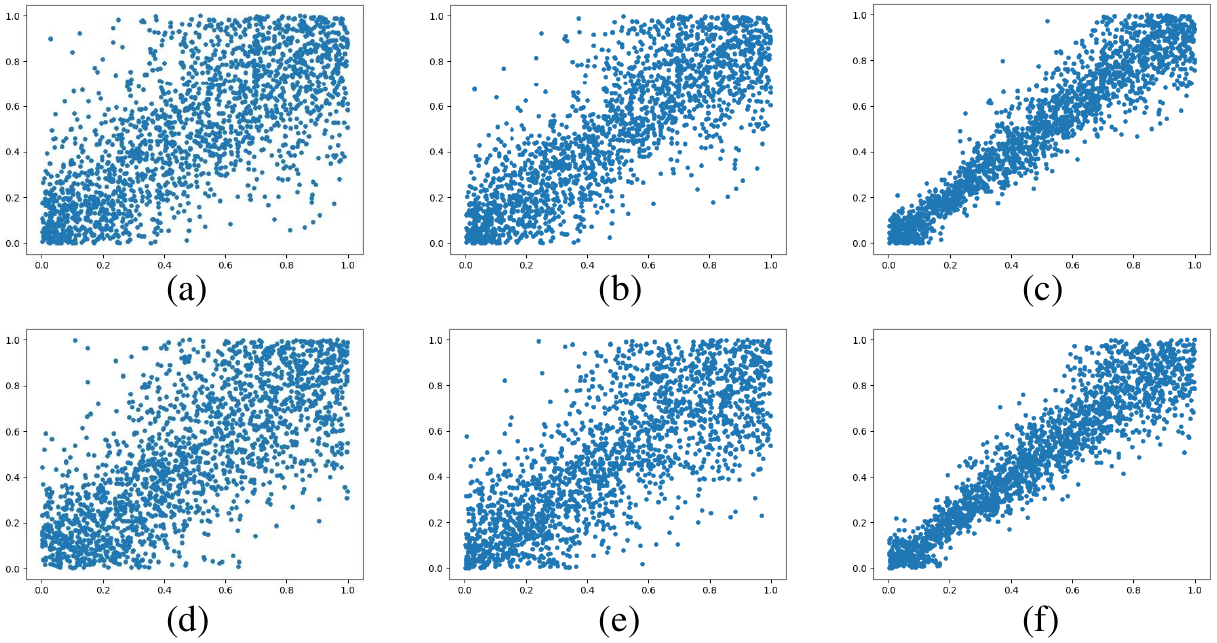}
    \caption{Inner Copulas Learned by HACsurv. (a), (b) and (c) are the copulas learned in the first stage. (d), (e) and (f) are the corresponding inner copulas learned in the second stage using the Re-generation Trick}
    \label{fig:innercopulas}
\end{figure}
\subsection{Additional Qualitative Results}

As shown in Figure~\ref{fig:innercopulas}, the inner copulas learned using the Re-generation trick proposed in this paper are very close to the copulas learned in the first stage. This confirms that HACSurv can effectively learn the inner copulas of HACs.

\begin{table}
\centering
\caption{Model Performance Comparison on Synthetic Dataset.}
\label{tab:synthetic_performance}
\footnotesize 
\setlength{\tabcolsep}{4pt} % 调整列间距
\begin{tabular}{lcccccc}
\toprule
Model & \multicolumn{2}{c}{Risk 1} & \multicolumn{2}{c}{Risk 2} & \multicolumn{2}{c}{Risk 3} \\
\cmidrule(r){2-3} \cmidrule(r){4-5} \cmidrule(r){6-7}
& \(C^{td}\)-index & IBS & \(C^{td}\)-index & IBS & \(C^{td}\)-index & IBS \\
\midrule
DeepHit       & $0.548 \pm 0.011$          & $0.196 \pm 0.003$          & $0.618 \pm 0.008$          & $0.153 \pm 0.004$          & $0.668 \pm 0.006$          & $0.105 \pm 0.002$          \\
DSM           & $0.552 \pm 0.009$          & $0.080 \pm 0.002$ & $0.593 \pm 0.015$          & $0.085 \pm 0.004$          & $0.665 \pm 0.004$          & $0.092 \pm 0.003$ \\
DeSurv        & $0.534 \pm 0.010$          & $0.196 \pm 0.002$          & $0.648 \pm 0.006$          & $0.131 \pm 0.003$          & $0.670 \pm 0.004$          & $0.102 \pm 0.002$          \\
NeuralFG      & $0.510 \pm 0.004$          & $0.211 \pm 0.007$          & $0.470 \pm 0.048$          & $0.194 \pm 0.039$          & $0.668 \pm 0.004$          & $0.115 \pm 0.003$          \\
HACSurv (I)   & $0.590 \pm 0.005$          & $\textbf{0.069} \pm 0.001$ & $0.644 \pm 0.008$          & $\textbf{0.073} \pm 0.002$ & $\textbf{0.671} \pm 0.003$ & $\textbf{0.090} \pm 0.003$ \\
HACSurv (S)   & $0.590 \pm 0.006$          & $0.085 \pm 0.003$          & $0.650 \pm 0.008$          & $0.085 \pm 0.006$          & $0.669 \pm 0.004$          & $0.095 \pm 0.004$          \\
HACSurv       & $\textbf{0.598} \pm 0.011$ & $0.088 \pm 0.011$          & $\textbf{0.659} \pm 0.007$ & $0.085 \pm 0.011$          & $\textbf{0.671} \pm 0.004$ & $0.126 \pm 0.003$          \\
\bottomrule
\end{tabular}
\end{table}

\begin{table}[ht]
\centering
\caption{Model Performance Comparison on Framingham and SEER Datasets.}
\label{tab:combined_performance}
\footnotesize  % Smaller font size
\setlength{\tabcolsep}{4pt} % Adjust column spacing
\begin{tabular}{llcccc}
\toprule
\multirow{2}{*}{Dataset} & \multirow{2}{*}{Model} & \multicolumn{2}{c}{Risk 1} & \multicolumn{2}{c}{Risk 2} \\
\cmidrule(r){3-4} \cmidrule(r){5-6}
& & \(C^{td}\)-index & IBS & \(C^{td}\)-index & IBS \\
\midrule
\multirow{7}{*}{Framingham}
& DeepHit        & $0.668 \pm \text{\scriptsize{0.028}}$ & $0.079 \pm \text{\scriptsize{0.003}}$ & $0.730 \pm \text{\scriptsize{0.011}}$ & $0.099 \pm \text{\scriptsize{0.003}}$ \\
& DSM            & $0.708 \pm \text{\scriptsize{0.016}}$ & $0.078 \pm \text{\scriptsize{0.003}}$ & $0.761 \pm \text{\scriptsize{0.017}}$ & $0.097 \pm \text{\scriptsize{0.004}}$ \\
& DeSurv         & $0.599 \pm \text{\scriptsize{0.040}}$ & $0.092 \pm \text{\scriptsize{0.006}}$ & $0.748 \pm \text{\scriptsize{0.004}}$ & $0.115 \pm \text{\scriptsize{0.001}}$ \\
& NeuralFG       & $0.719 \pm \text{\scriptsize{0.015}}$ & $0.083 \pm \text{\scriptsize{0.003}}$ & $0.774 \pm \text{\scriptsize{0.005}}$ & $0.098 \pm \text{\scriptsize{0.003}}$ \\
& HACSurv (I)    & $\textbf{0.720} \pm \text{\scriptsize{0.011}}$ & $0.081 \pm \text{\scriptsize{0.003}}$ & $\textbf{0.776} \pm \text{\scriptsize{0.007}}$ & $0.103 \pm \text{\scriptsize{0.006}}$ \\
& HACSurv (S)    & $0.718 \pm \text{\scriptsize{0.017}}$ & $0.081 \pm \text{\scriptsize{0.004}}$ & $0.773 \pm \text{\scriptsize{0.008}}$ & $0.101 \pm \text{\scriptsize{0.008}}$ \\
& HACSurv        & $\textbf{0.720} \pm \text{\scriptsize{0.015}}$ & $\textbf{0.076} \pm \text{\scriptsize{0.002}}$ & $0.774 \pm \text{\scriptsize{0.008}}$ & $\textbf{0.094} \pm \text{\scriptsize{0.004}}$ \\
\midrule
\multirow{7}{*}{SEER}
& DeepHit        & $0.746 \pm \text{\scriptsize{0.000}}$ & $\textbf{0.043} \pm \text{\scriptsize{0.000}}$ & $0.835 \pm \text{\scriptsize{0.000}}$ & $0.104 \pm \text{\scriptsize{0.001}}$ \\
& DSM            & $0.735 \pm \text{\scriptsize{0.009}}$ & $0.044 \pm \text{\scriptsize{0.000}}$ & $0.827 \pm \text{\scriptsize{0.002}}$ & $0.093 \pm \text{\scriptsize{0.000}}$ \\
& DeSurv         & $0.739 \pm \text{\scriptsize{0.003}}$ & $0.045 \pm \text{\scriptsize{0.000}}$ & $0.834 \pm \text{\scriptsize{0.007}}$ & $\textbf{0.090} \pm \text{\scriptsize{0.002}}$ \\
& NeuralFG       & $0.725 \pm \text{\scriptsize{0.005}}$ & $0.045 \pm \text{\scriptsize{0.000}}$ & $0.833 \pm \text{\scriptsize{0.002}}$ & $0.091 \pm \text{\scriptsize{0.001}}$ \\
& HACSurv (I)    & $\textbf{0.749} \pm \text{\scriptsize{0.005}}$ & $\textbf{0.043} \pm \text{\scriptsize{0.000}}$ & $0.846 \pm \text{\scriptsize{0.001}}$ & $0.091 \pm \text{\scriptsize{0.001}}$ \\
& HACSurv (S)    & $\textbf{0.749} \pm \text{\scriptsize{0.004}}$ & $\textbf{0.043} \pm \text{\scriptsize{0.001}}$ & $0.846 \pm \text{\scriptsize{0.001}}$ & $0.091 \pm \text{\scriptsize{0.001}}$ \\
& HACSurv        & $\textbf{0.749} \pm \text{\scriptsize{0.003}}$ & $\textbf{0.043} \pm \text{\scriptsize{0.000}}$ & $\textbf{0.847} \pm \text{\scriptsize{0.002}}$ & $0.091 \pm \text{\scriptsize{0.000}}$ \\
\bottomrule
\end{tabular}
\end{table}

\begin{table}[ht]
\centering 
\caption{Model Performance Comparison on MIMIC-III Dataset.}
\label{tab:MIMIC_performance}
\footnotesize
\setlength{\tabcolsep}{5pt} % Adjust column spacing
\begin{tabular}{lcccccc}
\toprule
\textbf{Metric} & \textbf{Model} & \textbf{Risk 1} & \textbf{Risk 2} & \textbf{Risk 3} & \textbf{Risk 4} & \textbf{Risk 5} \\
\midrule
\multirow{7}{*}{\(C^{td}\)-index} 
& DeepHit                     & $\mathbf{0.759} \pm 0.006$ & $0.844 \pm 0.018$ & $0.693 \pm 0.018$ & $0.768 \pm 0.013$ & $0.760 \pm 0.017$ \\
& DSM                         & $0.724 \pm 0.013$ & $0.829 \pm 0.016$ & $0.693 \pm 0.021$ & $0.768 \pm 0.016$ & $0.745 \pm 0.026$ \\
& DeSurv                      & $0.751 \pm 0.004$ & $\mathbf{0.859} \pm 0.028$ & $0.696 \pm 0.017$ & $0.742 \pm 0.035$ & $0.729 \pm 0.023$ \\
& NeuralFG                    & $\mathbf{0.759} \pm 0.010$ & $0.845 \pm 0.031$ & $\mathbf{0.711} \pm 0.015$ & $0.757 \pm 0.031$ & $0.730 \pm 0.042$ \\
& HACSurv (I)                 & $0.756 \pm 0.010$ & $0.851 \pm 0.026$ & $0.695 \pm 0.017$ & $0.764 \pm 0.020$ & $0.752 \pm 0.018$ \\
& HACSurv (S)                 & $0.752 \pm 0.010$ & $0.848 \pm 0.017$ & $0.680 \pm 0.020$ & $0.736 \pm 0.019$ & $\mathbf{0.786} \pm 0.015$ \\
& HACSurv                     & $0.752 \pm 0.008$ & $0.856 \pm 0.021$ & $0.690 \pm 0.018$ & $\mathbf{0.777} \pm 0.012$ & $0.767 \pm 0.034$ \\
\midrule
\multirow{7}{*}{IBS}
& DeepHit                     & $0.219 \pm 0.008$ & $0.232 \pm 0.088$ & $0.324 \pm 0.073$ & $0.269 \pm 0.127$ & $0.275 \pm 0.120$ \\
& DSM                         & $0.146 \pm 0.025$ & $0.221 \pm 0.104$ & $0.228 \pm 0.020$ & $0.255 \pm 0.128$ & $0.279 \pm 0.123$ \\
& DeSurv                      & $0.188 \pm 0.027$ & $0.245 \pm 0.096$ & $0.332 \pm 0.053$ & $0.281 \pm 0.124$ & $0.295 \pm 0.128$ \\
& NeuralFG                    & $0.255 \pm 0.029$ & $0.238 \pm 0.102$ & $0.355 \pm 0.092$ & $0.278 \pm 0.143$ & $0.253 \pm 0.120$ \\
& HACSurv (I)                 & $0.138 \pm 0.034$ & $0.130 \pm 0.049$ & $0.204 \pm 0.012$ & $0.205 \pm 0.082$ & $0.169 \pm 0.067$ \\
& HACSurv (S)                 & $\mathbf{0.103} \pm 0.007$ & $\mathbf{0.095} \pm 0.006$ & $\mathbf{0.183} \pm 0.044$ & $0.189 \pm 0.045$ & $\mathbf{0.084} \pm 0.004$ \\
& HACSurv                     & $0.137 \pm 0.057$ & $0.105 \pm 0.059$ & $0.237 \pm 0.095$ & $\mathbf{0.151} \pm 0.064$ & $0.184 \pm 0.081$ \\
\bottomrule
\end{tabular}
\end{table}

\subsection{Risk-Specific Results}
According to Table~\ref{tab:synthetic_performance}, HACSurv achieves the best c-index metric across all risks on the synthetic dataset. HACSurv (Independent) obtained the best IBS, indicating that IBS is not a strictly proper evaluation metric. According to Table~\ref{tab:combined_performance}, NeuralFG, which also uses sumo-net as the survival marginal, achieves competitive results with HACSurv (Independent) and HACSurv (Symmetry). However, their results are inferior to HACSurv because they fail to accurately capture the dependency structure among the two risks and censoring. For the SEER dataset, we found that the learned dependency was relatively weak, hence the three versions of HACSurv performed similarly. As shown in Table~\ref{tab:MIMIC_performance} and the overall results in the main text, HACSurv achieves the best overall C-index. HACSurv (Symmetry) obtained the best IBS. Modeling the dependency structure of survival data with six dimensions (including censoring) is challenging. HACSurv (Symmetry), using an Archimedean copula to capture their dependencies, surpasses all existing methods in predicting survival outcomes.

\section{EXPERIMENTAL DETAILS}

All experiments are conducted with a single NVIDIA RTX3090 GPU. We utilize the AdamW optimizer~\cite{loshchilov2018AdamW} for training. HACSurv employs the same network architecture across all datasets. Both the shared embedding network and the cause-specific embedding networks consist of two layers with 100 neurons. The monotonic neural networks are composed of three fully connected layers, each containing 100 neurons. For the inner and outer generators of HAC, we use the same network architecture as \cite{ng2021generative}. For more details, please refer to our code.

The split between the training and testing sets for all datasets is 8:2. 20\% of the data in the training set is used as a validation set. Experiments across all datasets are conducted in five batches, using 41, 42, 43, 44, and 45 as the random seeds for splitting the datasets. For all comparative methods, a grid search is conducted to select hyperparameters. To ensure a fair comparison, all methods use the same time grid for computing metrics during testing. 

For the implementation of the compared algorithms, we use the \texttt{PyCox} library \cite{Kvamme2019}, which implements DeepHit. The code for NeuralFG, DeSurv, and DSM can be found at \url{https://github.com/Jeanselme/NeuralFineGray}.

\end{document}